\documentclass[letterpaper]{article} 
\usepackage{aaai23}  
\usepackage[ruled,vlined]{algorithm2e}
\usepackage{times}  
\usepackage{helvet}  
\usepackage{courier}  
\usepackage[hyphens]{url}  
\usepackage{graphicx} 
\usepackage{natbib}  
\usepackage{caption} 
\usepackage{lipsum}
\usepackage{color, colortbl}
\usepackage{setspace}
\usepackage{overpic}
\usepackage{hyperref}
\usepackage{amsmath,amssymb}
\usepackage{cleveref}
\newcommand\norm[1]{\lVert#1\rVert}
\usepackage{booktabs, wrapfig}
\usepackage{adjustbox}
\usepackage{multirow}
\definecolor{LightCyan}{rgb}{0.88,1,1}
\newcommand{\ourmethod}{\texttt{TRIDENT}}
\newcommand{\ourmodule}{\texttt{AttFEX}}
\usepackage[table,xcdraw]{xcolor}
\usepackage{color, colortbl}
\usepackage{academicons}
\definecolor{orcidlogocol}{HTML}{A6CE39}
\usepackage{newfloat}
\usepackage{listings}
\DeclareCaptionStyle{ruled}{labelfont=normalfont,labelsep=colon,strut=off} 
\title{Transductive Decoupled Variational Inference for Few-Shot Classification}
\author{
    Anuj Singh \textsuperscript{\rm 1},\quad
    Hadi Jamali-Rad \textsuperscript{\rm 1, 2}
}
\affiliations{
    {\textsuperscript{\rm 1}Delft University of Technology, The Netherlands}, \\
    {\textsuperscript{\rm 2}Shell Global Solutions International B.V., Amsterdam, The Netherlands}

}
\begin{document}
\maketitle

\begin{abstract}
The versatility to learn from a handful of samples is the hallmark of human intelligence. Few-shot learning is an endeavour to transcend this capability down to machines. Inspired by the promise and power of probabilistic deep learning, we propose a novel variational inference network for few-shot classification (coined as \ourmethod) to decouple the representation of an image into \emph{semantic} and \emph{label} latent variables, and simultaneously infer them in an intertwined fashion. To induce \emph{task-awareness}, as part of the inference mechanics of \ourmethod, we exploit information across both query and support images of a few-shot task using a novel built-in attention-based transductive feature extraction module (we call \ourmodule). Our extensive experimental results corroborate the efficacy of \ourmethod{} and demonstrate that, using the simplest of backbones, it sets a new state-of-the-art in the most commonly adopted datasets \emph{mini}ImageNet and \emph{tiered}ImageNet (offering up to $4\%$ and $5\%$ improvements, respectively), as well as for the recent challenging cross-domain \textit{mini}Imagenet $\rightarrow$ CUB scenario offering a significant margin (up to $20\%$ improvement) beyond the best existing cross-domain baselines.\footnote{Code and experimentation can be found in our GitHub repository: \url{https://github.com/anujinho/trident}}
\end{abstract}
\section{Introduction}
Deep learning algorithms are usually data hungry and require massive amounts of training data to reach a satisfactory level of performance on any task. To tackle this limitation, few-shot classification aims to learn to classify images from various unseen tasks in a data-deficient setting. In this exciting space, \textit{metric learning} proposes to learn a shared feature extractor to embed the samples into a metric space of class prototypes \cite{Sung_2018_CVPR, NIPS2016_90e13578, NIPS2017_cb8da676, wang2019simpleshot, DBLP:conf/eccv/LiuSQ20, Bateni2020_SimpleCNAPS}. Due to limited data per class, these prototypes suffer from sample-bias and fail to efficiently represent class characteristics. Furthermore, sharing a feature extractor across tasks implies that the discriminative information learnt from the seen classes are equally effective on any arbitrary unseen classes, which is not true in most cases. \emph{Task-aware} few-shot learning approaches \cite{Bateni2022_TransductiveCNAPS, ye2020fewshot} address these limitations by exploiting information hidden in the unlabeled data. As a result, the model learns task-specific embeddings by aligning the features of the labelled and unlabelled task instances for optimal distance metric based label assignment. Since the alignment of these embeddings is still subject to the relevance of the characteristics captured by the shared feature extractors, task-aware methods sometimes fail to extract meaningful representations particularly relevant to classification. \emph{Probabilistic} methods address sample-bias by relaxing the need to find point estimates to approximate data-dependent distributions of either high-dimensional model weights \cite{vampire2019, ravi2018amortized, gordon2018versa, Hu2020Empirical} or lower-dimensional class prototypes \cite{pmlr-v130-sun21a, 9010263}. However, inferring a high-dimensional posterior of model parameters is inefficient in low-data regimes and estimating distributions of class prototypes involves using hand-crafted non-parametric aggregation techniques which may not be well suited for every unseen task.
\begin{figure*}[t!]
    \centering
    \begin{overpic}[abs, unit=1cm, width=0.85\textwidth]{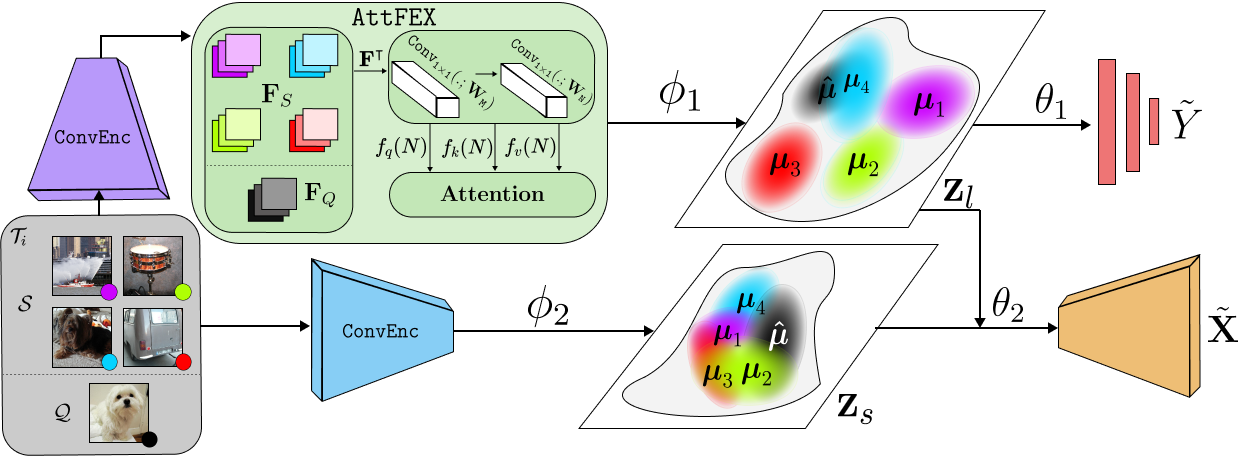}
    \end{overpic}
    \caption{\small High-level process flow of \ourmethod. Inferred label latent variable $\textbf{z}_l$ contains class-characterizing information, as is reflected by better separation of the distributions when compared to their semantic latent counterparts $\textbf{z}_s$. \texttt{AttFEX} module generates \textit{task-aware} feature maps by exploiting information from both support and query images, which compensates for the lack of label vectors $Y$ in inferring $\textbf{z}_l$.}
    
    \label{fig:intuition}
\end{figure*}

Although fit for purpose, all these approaches seem to overlook an important perspective. An image is composed of different attributes such as style, design, and context, which are not necessarily relevant discriminative characteristics for classification. Here, we refer to these attributes as \emph{semantic} information. On the other hand, other class-characterizing attributes (such as wings of a bird, trunk of an elephant, hump on a camel's back) are critical for classification, irrespective of context. We refer to such attributes as \emph{label} information. Typically, contextual information is majorly governed by semantic attributes, whereas the label characteristics are subtly embedded throughout an image. In other words, semantic information can be predominantly present across an image, whereas \emph{attending} to subtle label information determines how effective a classification algorithm would be. Thus, we argue that attention to label-specific information should be ingrained into the mechanics of the classifier, decoupling it from semantic information. This becomes even more important in a few-shot setting where the network has to quickly learn from little data. Building upon this idea, we propose \textbf{tr}ansductive variational \textbf{i}nference of \textbf{de}coupled late\textbf{nt} variables (coined as \texttt{TRIDENT}), to simultaneously infer decoupled label and semantic information using two intertwined variational networks. To induce task-awareness while constructing the variational inference mechanics of \ourmethod, we introduce a novel \textbf{at}ention-based \textbf{t}ransductive \textbf{f}eature \textbf{ex}traction module (we call \ourmodule) which further enhances the discriminative power of the inferred label attributes. This way \ourmethod{} infers distributions instead of point estimates and injects a handcrafted inductive-bias into the network to guide the classification process. Our main contributions can be summarized as: 
\begin{enumerate}
    \item We propose \texttt{TRIDENT}, a variational inference network to simultaneously infer two salient \emph{decoupled} attributes of an image (\emph{label} and \emph{semantic}), by inferring these two using two intertwined variational sub-networks (Fig.~\ref{fig:intuition}). 
    \item We introduce an attention-based transductive feature extraction module, \texttt{AttFEX}, to enable \ourmethod{} see through and compare all images within a task, inducing task-cognizance in the inference of label information.
    \item We perform extensive evaluations to demonstrate that \ourmethod{} sets a new state-of-the-art by outperforming all existing baselines on the most commonly adopted datasets \emph{mini}Imagenet and \emph{tiered}Imagenet (up to $4\%$ and $5\%$), as well as for the challenging cross-domain scenario of \textit{mini}Imagenet $\rightarrow$ CUB (up to $20\%$ improvement).
\end{enumerate}
\section{Related Work}
\textbf{Metric-based learning.}
This body of work revolves around mapping input samples into a lower-dimensional embedding space and then classifying the unlabelled samples based on a distance or similarity metric. By parameterizing these mappings with neural networks and using differentiable similarity metrics for classification, these networks can be trained in an episodic manner \cite{NIPS2016_90e13578} to perform few-shot classification. Prototypical Nets \cite{NIPS2017_cb8da676}, Simple Shot \cite{wang2019simpleshot}, Relation Networks \cite{Sung_2018_CVPR}, Matching Networks \cite{NIPS2016_90e13578} variants of Graph Neural Nets \cite{garcia2018fewshot, Yang2020DPGNDP}, Simple CNAPS \cite{Bateni2020_SimpleCNAPS}, are a few examples of seminal ideas in this space. 

\noindent\textbf{Transductive Feature-Extraction and Inference.}
Transductive feature extraction or task-aware learning is a variant of the metric-learning with an adaptation mechanism that \emph{aligns} support and query feature vectors in the embedding space for better representation of task-specific discriminative information. This not only improves the discriminative ability of classifiers across tasks, but also alleviates the problem of overfitting on limited support set since information from the query set is also used for extracting features of images in a task. CNAPS \cite{NEURIPS2019_1138d90e},  Transductive-CNAPS \cite{Bateni2022_TransductiveCNAPS}, FEAT \cite{ye2020fewshot}, Assoc-Align \cite{afrasiyabi2020associative}, TPMN \cite{wu2021task} and CTM \cite{li2019ctm} are prime examples of such methods. Next to transduction for task-aware feature extraction, there are methods that use \emph{transductive inference} to classify all the query samples at once by jointly assigning them labels, as opposed to their inductive counterparts where prediction is done on the samples one at a time. This is either done by iteratively propagating labels from the support to the query samples or by fine-tuning a pre-trained backbone using an additional entropy loss on all query samples, which encourages confident class predictions at query samples. TPN \cite{liu2019fewTPN}, Ent-Min \cite{Dhillon2020A}, TIM \cite{NEURIPS2020_196f5641}, Transductive-CNAPS \cite{Bateni2022_TransductiveCNAPS}, LaplacianShot \cite{DBLP:conf/icml/ZikoDGA20}, DPGN \cite{Yang2020DPGNDP} and ReRank \cite{shen2021reranking} are a few notable examples in this space that usually report state-of-the-art results in certain few-shot classification settings \cite{liu2019fewTPN}. 

\noindent\textbf{Optimization-based meta-learning.} These methods search for model parameters that are sensitive to task objective functions for fast gradient-based adaptation to new tasks. MAML \cite{pmlr-v70-finn17a}, its variants \cite{NEURIPS2019_072b030b, nichol2018firstorder} and SNAIL \cite{mishra2018a} are a few prominent examples while LEO \cite{rusu2018metalearning} efficiently meta-updates its parameters in a lower dimensional latent space. 

\noindent\textbf{Probabilistic learning.} The estimated parameters of typical gradient-based meta-learning methods discussed earlier \cite{pmlr-v70-finn17a, rusu2018metalearning, mishra2018a, nichol2018firstorder, NEURIPS2019_072b030b}, have high variance due to the small task sample size. To deal with this, a natural extension is to model the uncertainty by treating these parameters as latent variables in a Bayesian framework as proposed in Neural Statistician \cite{Edwards2017TowardsAN}, PLATIPUS \cite{platipus}, VAMPIRE \cite{vampire2019}, ABML \cite{ravi2018amortized}, VERSA \cite{gordon2018versa}, SIB \cite{Hu2020Empirical}, SAMOVAR \cite{pmlr-v119-iakovleva20a}. 
Methods like ABPML \cite{pmlr-v130-sun21a} and VariationalFSL \cite{9010263} infer latent variables of class prototypes to perform classification and avoid inferring high-dimensional model parameters. ABPML \cite{pmlr-v130-sun21a} and VariationalFSL \cite{9010263} are the closest to our approach. In contrast to these two methods, we avoid hand-crafting class-level aggregations, as well as we enhance variational inference by incorporating an inductive bias through decoupling of label and semantic information. 
\section{Problem Definition}
Consider a labelled dataset $\mathcal{D} = \{(\textbf{x}_i, y_i)\, |\, i \in [1,N^\prime] \}$ of images $\textbf{x}_i$ and class labels $y_i$. This dataset $\mathcal{D}$ is divided into three disjoint subsets: $\mathcal{D} = \{\mathcal{D}^{tr}\, \cup\, \mathcal{D}^{val}\, \cup\, \mathcal{D}^{test}\}$, respectively, referring to the training, validation, and test subsets. The validation dataset $\mathcal{D}^{val}$ is used for model selection and the testing dataset $\mathcal{D}^{test}$ for final evaluation. Following standard few-shot classification settings \cite{NIPS2016_90e13578, Sung_2018_CVPR, NIPS2017_cb8da676}, we use episodic training on a set of tasks $\mathcal{T}_i \sim p(\mathcal{T})$. The tasks are constructed by drawing $K$ random samples from $N$ different classes, which we denote as an ($N$-way, $K$-shot) task. Concretely, each task $\mathcal{T}_i$ is composed of a $\textit{support}$ and a $\textit{query}$ set. The support set $\mathcal{S} = \{(\textbf{x}_{kn}^{S}, y_{kn}^{S})\, |\, k \in [1,K], n \in [1,N]\}$ contains $K$ samples per class and the query set $\mathcal{Q} = \{(\textbf{x}_{kn}^{Q}, y_{kn}^{Q})\, |\, k \in [1,Q], n \in [1,N]\}$ contains $Q$ samples per class. For a given task, the $NQ$ query and $NK$ support images are mutually exclusive to assess the generalization performance. 

\section{The Proposed Method: \texttt{TRIDENT}}
Let us start with the high-level idea. The proposed approach is devised to learn meaningful representations that capture two pivotal characteristics of an image by modelling them as separate latent variables: (i) ${\bf z}_s$ representing \textit{semantics}, and (ii) ${\bf z}_l$ embodying class \textit{labels}. Inferring these two latent variables simultaneously allows $\textbf{z}_l$ to learn meaningful distributions of class-discriminating characteristics \textit{decoupled} from semantic features represented by $\textbf{z}_s$. We argue that learning $\textbf{z}_l$ as the sole latent variable for classification results in capturing a mixture of true label and other semantic information. This in turn can lead to sub-optimal classification performance, especially in a few-shot setting where the information per class is scarce and the network has to adapt and generalize quickly. By inferring decoupled label and semantics latent variables, we inject a handcrafted inductive-bias that incorporates only relevant characteristics, and thus, ameliorates the network's classification performance.  
\begin{figure}[t!]
\centering
\includegraphics[width=\columnwidth]{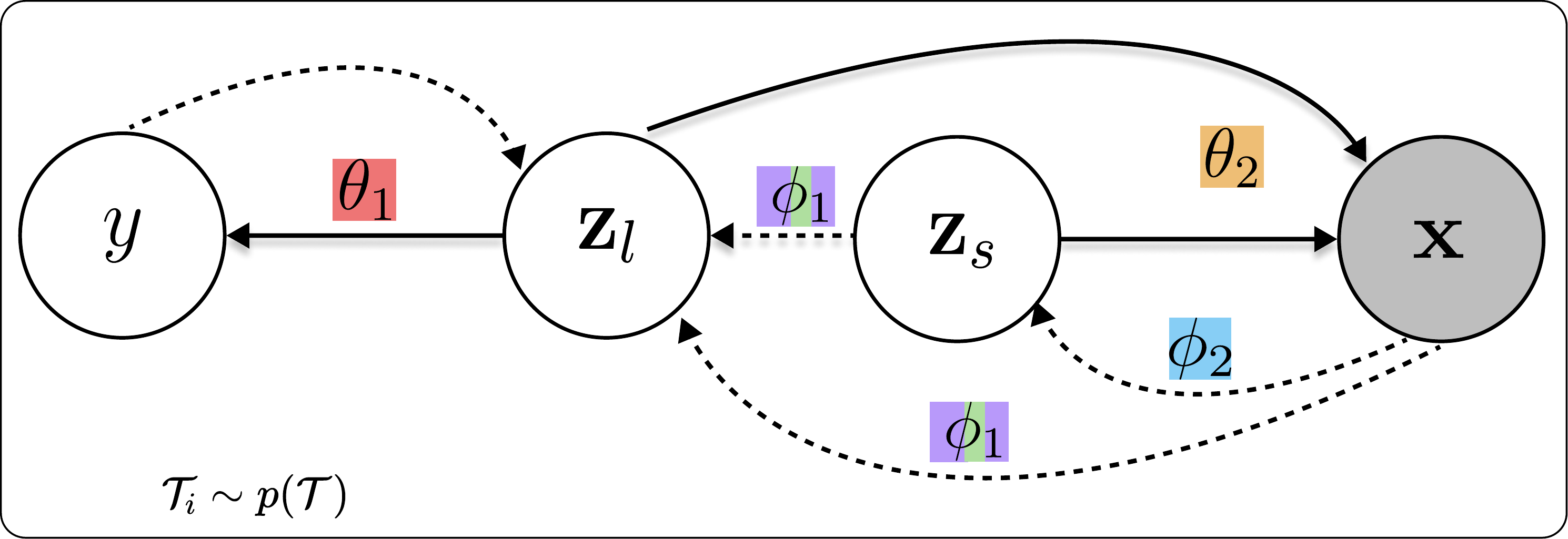}
\caption{\small Generative Model of \texttt{TRIDENT}. Dotted lines indicate variational inference and solid lines refer to generative processes. The inference and generative parameters are color coded to correspond to their respective architectures indicated in Fig.\ref{fig:intuition} and Fig.\ref{fig:nn}.}
\label{fig:genmod}
\end{figure}
%
\subsection{Generative Process}
The directed graphical model in Fig.~\ref{fig:genmod} illustrates the common underlying generative process $p$ such that $p_i = p(\textbf{x}_i, y_i\,\vert\,\textbf{z}_{li}, \textbf{z}_{si})$. For the sake of brevity, in the following we drop the sample index $i$ as we always refer to terms associated with a single data sample. We work on the logical premise that the label latent variable $\textbf{z}_l$ is responsible for generating class label as well as for image reconstruction, whereas the semantic latent variable $\textbf{z}_s$ is only responsible for image reconstruction (solid lines in the figure). Formally, the data is explained by the generative processes: $p_{\theta_1}(y\,\vert\,\textbf{z}_l) = \textup{Cat}(y\,\vert\,\textbf{z}_l)$ and $p_{\theta_2}(\textbf{x}\,\vert\,\textbf{z}_l, \textbf{z}_s) = g_{\theta_2}(\textbf{x};\textbf{z}_l, \textbf{z}_s)$, where Cat(.) refers to a multinomial distribution and $g_{\theta_2}(\textbf{x};\textbf{z}_l, \textbf{z}_s)$ is a suitable likelihood function such as a Gaussian or Bernoulli distribution.
The likelihoods of both these generative processes are parameterized using deep neural networks and the priors of the latent variables are chosen to be standard multivariate Gaussian distributions \cite{kingma2014autoencoding, NIPS2014_d523773c}:
$p(\textbf{z}_{s}) = \mathcal{N}(\textbf{z}_{s}\,\vert\,\textbf{0}, \textbf{I})$ and $p(\textbf{z}_{l}) = \mathcal{N}(\textbf{z}_{l}\,\vert\,\textbf{0}, \textbf{I})$.
\subsection{Variational Inference of Decoupled ${\bf Z}_l$ and ${\bf Z}_s$}
Computing exact posterior distributions is intractable due to high dimensionality and non-linearity of the deep neural network parameter space. Following \cite{kingma2014autoencoding, NIPS2014_d523773c}, we instead construct an approximate posterior over the latent variables by introducing a fixed-form distribution $q(\textbf{z}_l, \textbf{z}_s\,|\,\textbf{x}, y)$ parameterized by $\phi$. By using $q_{\phi}(.)$ as an inference network, the inference is rendered tractable, scalable and amortized since $\phi$ now acts as the global variational parameter. We assume $q_{\phi}$ has a factorized form $q_{\phi}\left(\textbf{z}_{s}, \textbf{z}_{l} \mid \textbf{x}, y\right) = q_{\phi_1}\left(\textbf{z}_{l} \mid \textbf{x}, \textbf{z}_{s}\right) q_{\phi_2}\left(\textbf{z}_{s} \mid \textbf{x}\right)$, where $q_{\phi_1}(.), q_{\phi_2}(.)$ are assumed to be multivariate Gaussian distributions. As is also depicted in Fig.~\ref{fig:genmod}, we use ${\bf z}_s$ as input to $q_{\phi_1}(.)$ to infer ${\bf z}_l$ because of their conditional dependence given ${\bf x}$. This way we forge a path to allow \emph{necessary} semantic latent information flow through the label inference network. On the other hand, the opposite direction (using ${\bf z}_l$ to infer ${\bf z}_s$) is unnecessary, because label information does not directly contribute to the extraction of semantic features. We will further reflect on this design choice in the next subsection. Neural networks are then used to parameterize both inference networks as:
\begin{equation}
\small
\begin{split}
&q_{\phi_{2}}\left(\textbf{z}_{s} \,\vert\, \textbf{x}\right) = \mathcal{N}\left(\textbf{z}_{s} \,\vert\, \boldsymbol{\mu}_{\phi_{2}}(\textbf{x}), diag(\boldsymbol{\sigma}^{2}_{\phi_{2}}(\textbf{x}))\right), \\
&q_{\phi_{1}}\left(\textbf{z}_{l} \,\vert\, \textbf{x}, \textbf{z}_{s}\right) = \mathcal{N}\left(\textbf{z}_{l} \,\vert\, \boldsymbol{\mu}_{\phi_{1}}(\textbf{x}, \textbf{z}_{s}), diag(\boldsymbol{\sigma}^{2}_{\phi_{1}}(\textbf{x}, \textbf{z}_{s}))\right). \label{eq:q1}
\end{split}
\end{equation}

To find the optimal \emph{approximate} posterior, we derive the evidence lower bound (ELBO) on the marginal likelihood of the data to form our objective function:
\begin{equation}
\small
\begin{split}
&p(\textbf{x}, y) =\iint p(\textbf{x}, y \,\vert\, \textbf{z}_{s}, \textbf{z}_{l}) \, p(\textbf{z}_{s,} \textbf{z}_{l}) \, d \textbf{z}_{s} \, d \textbf{z}_{l},  \\
&=\mathbb{E}_{q(\textbf{z}_{s}, \textbf{z}_{l} \,\vert\, x)}\left[\frac{p(\textbf{x} \,\vert\, \textbf{z}_{l}, \textbf{z}_{s}) p(y \,\vert\, \textbf{z}_{l}) p(\textbf{z}_{l}) p(\textbf{z}_{s})}{q(\textbf{z}_{l}, \textbf{z}_{s} \,\vert\, \textbf{x})}\right].  \\
&\ln p(\textbf{x}, y) \geqslant \mathbb{E}_{q(\textbf{z}_{s}, \textbf{z}_{l}\vert\textbf{x})}\left[\ln \left(\frac{p(\textbf{x} \,\vert\, \textbf{z}_{l}, \textbf{z}_{s}) p(y \,\vert\, \textbf{z}_{l}) p(\textbf{z}_{l}) p(\textbf{z}_{s})}{q(\textbf{z}_{s}, \textbf{z}_{l} \,\vert\, \textbf{x})}\right)\right],  \\
&=\mathbb{E}_{q_{\phi_2}}\left[\mathbb{E}_{q_{\phi_1}}\left[\ln \left(\frac{p(\textbf{x} \,\vert\, \textbf{z}_{s}, \textbf{z}_{l}) p(y \,\vert\, \textbf{z}_{l}) p(\textbf{z}_{s}) p(\textbf{z}_{l})}{q(\textbf{z}_{s} \,\vert\, \textbf{x}) q(\textbf{z}_{l} \,\vert\, \textbf{x}, \textbf{z}_{s})}\right)\right]\right].\nonumber
\end{split}
\end{equation}
Denoting $\Psi = \left(\theta_1, \theta_2, \phi_1, \phi_2\right)$, the ELBO can be given by
\begin{equation}
\small
\begin{split}
\label{eq:loss}
&\mathcal{L}(\Psi) = -\mathbb{E}_{q_{\phi_2}}\mathbb{E}_{q_{\phi_1}}\left[\ln p_{\theta_2}(\textbf{x} \,\vert\, \textbf{z}_{s}, \textbf{z}_{l}) + \ln p_{\theta_1}(y \,\vert\, \textbf{z}_{l})\right] + \\
& D_{KL}\big(q_{\phi_1}(\textbf{z}_{l} \,\vert\, \textbf{x}, \textbf{z}_{s}) \Vert\, p(\textbf{z}_{l})\big) + D_{KL}\big(q_{\phi_2}(\textbf{z}_{s} \,\vert\, \textbf{x}) \Vert\, p(\textbf{z}_{s})\big),
\end{split}
\end{equation}
where the second line follows the graphical model in Fig~\ref{fig:genmod}, and $\mathbb{E}(.)$ and $\ln(.)$ denote the expectation operator and the natural logarithm, respectively. We avoid computing biased gradients by following the re-parameterization trick from \cite{kingma2014autoencoding}. Assuming Gaussian distributions for the priors as well as the variational distributions allows us to compute the KL Divergences of $\textbf{z}_l$ and $\textbf{z}_s$ (last two terms in \eqref{eq:loss}) analytically \cite{kingma2014autoencoding}. By considering a multivariate Gaussian distribution and a multinomial distribution as the likelihood functions for $p_{\theta_2}\left(\textbf{x} \,\vert\, \textbf{z}_{s}, \textbf{z}_{l}\right)$ and $p_{\theta_1}\left(y \,\vert\, \textbf{z}_{l}\right)$, respectively, the negative log-likelihood of $\textbf{x}$ becomes the mean squared error (MSE) between the reconstructed images $\tilde{\textbf{x}}$ and the ground-truth images $\textbf{x}$ while the negative log-likelihood of $y$ becomes the cross-entropy between the actual labels $y$ and the predicted labels $\tilde{y}$. After working \eqref{eq:loss} out, we arrive at our overall objective function $\mathcal{L} = \mathcal{L}_R + \mathcal{L}_C$, where:
\begin{equation}
\small
\label{eq:nnloss}
\begin{split}
&\mathcal{L}_R = \alpha_1 \norm{\textbf{x} - \tilde{\textbf{x}}}^2 - KL(\mu_s, \sigma_s),  \\ 
&\mathcal{L}_C = -\alpha_2 \sum_{n=1}^{N} y_n\ln p_{\theta_1}(\tilde{y} = n \,\vert\, \textbf{z}_{l}) - KL(\mu_l, \sigma_l),
\end{split}
\end{equation}
where $KL(\mu, \sigma) = \frac{1}{2}\sum_{d=1}^{D} \big(1 + 2\ln(\sigma^d) - (\mu^d)^2 - (\sigma^d)^2\big)$, $D$ denotes the dimension of the latent space, $N$ is the total number of classes in an ($N$-way, $K$-shot) task, $\alpha_1, \alpha_2$ are constant scaling factors, $\mu_s$ and $\sigma^2_s$ denote the mean and variance vectors of semantic latent distribution, and $\mu_l$ and $\sigma^2_l$ denote the mean and variance vectors of label latent distribution. The hyper-parameters $\alpha_1, \alpha_2$ only scale the evidence lower-bound appropriately, since the reconstruction loss is in practice three orders of magnitude greater than the cross-entropy loss. As such, this scaling helps convergence but impacts the tightness of the ELBO slightly; nonetheless, \eqref{eq:loss} and \eqref{eq:nnloss} are still considered variational inference by consensus among the literature \cite{higgins2017betavae, joy2021capturing, pmlr-v97-mathieu19a, NEURIPS2018_b9228e09}. The loss is calculated for each given task on query and support sets separately; i.e., $\mathcal{L}^{g} = \mathcal{L}_R^{g} + \mathcal{L}_C^{g}$ with $g \in \{\mathcal{S}_i, \mathcal{Q}_i\}$. Note that in \eqref{eq:q1} we deliberately choose to exclude the label information $y$ as input to $q_{\phi_1}(.)$ to be able to exploit the associated generative network $p_{\theta_1}(y \,\vert\, \textbf{z}_{l})$ as a classifier. The consequence and the proposed solution to accommodate this design choice are discussed in the next subsection.
\section{\texttt{AttFEX} for Transductive Feature Extraction}
We first extract the feature maps of all images in the task using a convolutional block $\textbf{F} = \tt{ConvEnc}(\textbf{X})$ where $\textbf{X} \in \mathbb{R}^{N(K+Q) \times C \times W \times H}$, $\textbf{F} \in \mathbb{R}^{N(K+Q) \times C' \times W' \times H'}$. The feature map tensor $\textbf{F}$ is then transposed into $\textbf{F}' \in \mathbb{R}^{C' \times N(K+Q) \times W' \times H'}$ and fed into two consecutive $1\times1$ convolution blocks. This helps the network utilize information across corresponding pixels of all images in a task $\mathcal{T}_i$ which acts as a parametric comparison of classes. We leverage the fact that \texttt{ConvEnc} already extracts local pixel information by using larger kernels, and thus, use parameter-light $1\times1$ convolutions subsequently to focus only on individual pixels. Let $\textbf{F}'_{i}$ denote the $i^{th}$ channel (or feature map layer) out of total of $C'$ available and $\tt{ReLU}$ denote the rectified linear unit activation. The $1 \times 1$ convolution block ($\texttt{Conv}_{1\times1}$) is formulated as follows:
\begin{equation}
\begin{split}
&\textbf{M}_i = \texttt{ReLU}\big( \texttt{Conv}_{1\times1}(\textbf{F}'_{i}, \textbf{W}_M) \big), \forall i \in [1, C'];\,\\
&\textbf{N}_j = \texttt{ReLU}\big( \texttt{Conv}_{1\times1}(\textbf{M}_{j}, \textbf{W}_N) \big), \forall j \in [1, C'];\,
\end{split}
\end{equation}
where $\textbf{N} \in \mathbb{R}^{C' \times 32 \times W' \times H'}$ and $\textbf{W}_M \in \mathbb{R}^{64 \times N(K+Q) \times 1 \times 1}$, $\textbf{W}_N \in \mathbb{R}^{32 \times 64 \times 1 \times 1}$ denote the learnable weights. Next, we want to blend information across feature maps for which we use a self-attention mechanism \cite{NIPS2017_3f5ee243} across $\textbf{N}_j, \forall j \in [1, 32]$. To do so, we feed $\textbf{N}$ to query, key and value extraction networks $f_q(,;\textbf{W}_Q)$, $f_k(.; \textbf{W}_K)$, $f_v(.; \textbf{W}_V)$ which are also designed to be $1 \times 1$ convolutions as:
\begin{equation}
\begin{split}
\textbf{Q}_i = \texttt{ReLU}\left( \texttt{Conv}_{1\times1}(\textbf{N}_{i}, \textbf{W}_Q) \right), \quad \forall i \in [1, C'];\, \\
\textbf{K}_i = \texttt{ReLU}\left( \texttt{Conv}_{1\times1}(\textbf{N}_{i}, \textbf{W}_K) \right), \quad \forall i \in [1, C'];\, \\
\textbf{V}_i = \texttt{ReLU}\left( \texttt{Conv}_{1\times1}(\textbf{N}_{i}, \textbf{W}_V) \right), \quad \forall i \in [1, C'];\,
\end{split}
\end{equation}
\begin{figure}[t!]
\centering
\includegraphics[width=\columnwidth]{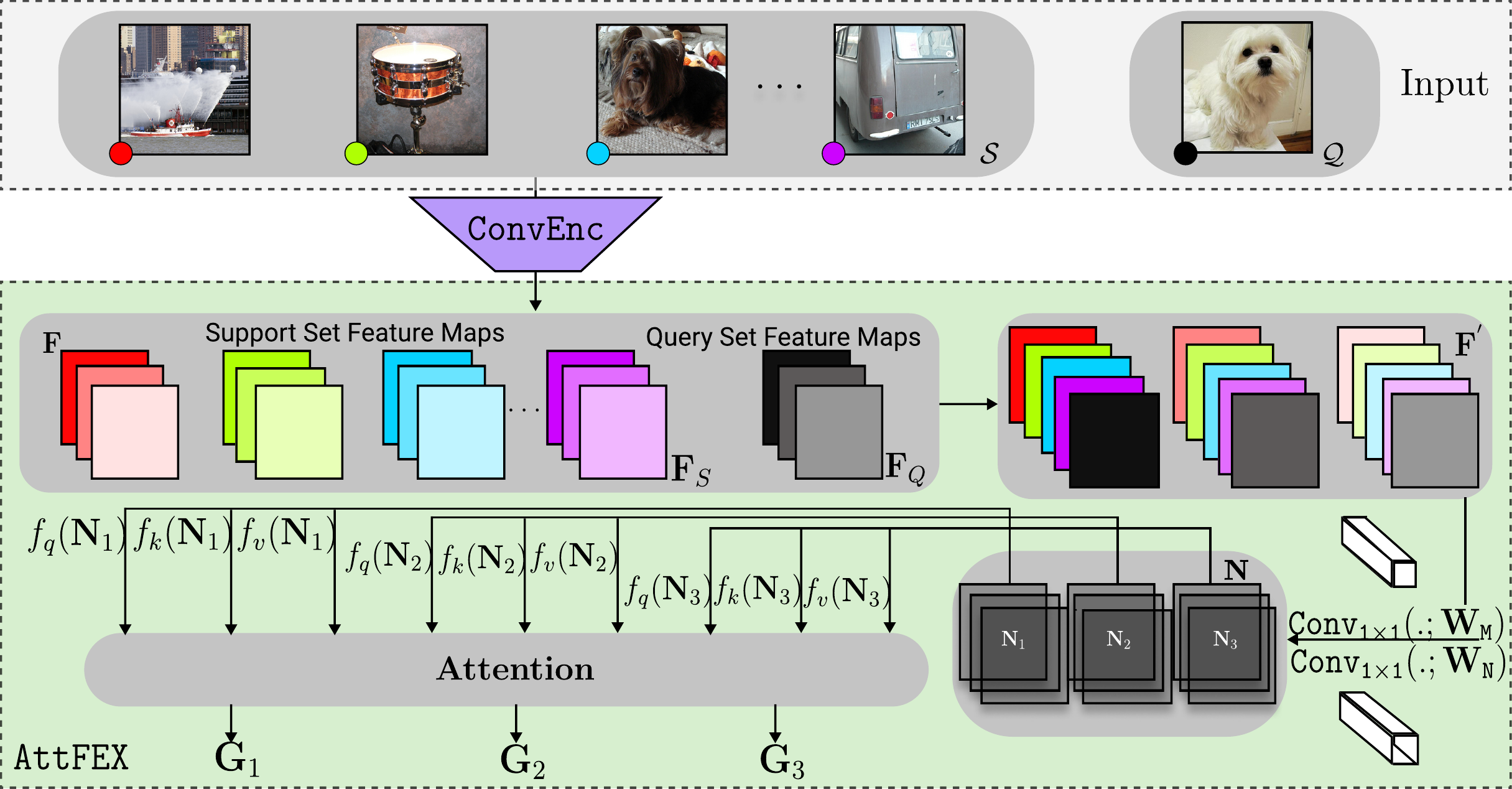}
\caption{\small \texttt{AttFEX} module depicting colors as images and shades as feature maps. We illustrate only 3 image feature maps and 3 channels instead of 32 for $\textbf{N}$, for the sake of simplicity.}
\label{fig:AttFEX}
\end{figure}
%
where $\textbf{W}_Q, \textbf{W}_K, \textbf{W}_V \in \mathbb{R}^{1 \times 32 \times 1 \times 1}$ are the learnable weights and $\textbf{Q}$, $\textbf{K}$, $\textbf{V}$ $\in \mathbb{R}^{C' \times 1 \times W' \times H'}$ are the query, key and value tensors. Next, each feature map $\textbf{N}_j$ is mapped to its output tensor $\textbf{G}_j$ by computing a weighted sum of the values, where each weight (within parentheses in \eqref{eq:attnetion_G}) measures the compatibility (or similarity) between the query and its corresponding key tensor using an inner-product: 
%
\begin{align}
\label{eq:attnetion_G}
\textbf{G}_i = \sum_{j=1}^{C'}\left({\frac{\exp{\left(\textbf{Q}_{i}\cdot\textbf{K}_{j}\right)}}{\sqrt{d_k} . \sum_{k=1}^{C'}{\exp{\left(\textbf{Q}_{i}\cdot\textbf{K}_{k}\right)}}}}\right) \textbf{V}_i,
\end{align}
where $d_k = W' \times H'$, and $\textbf{G}_i \in \mathbb{R}^{1 \times C' \times W' \times H'}$, $\forall i$. Finally, we transform the original feature maps $\mathbf{F}$ by applying a Hadamard product between the feature mask $\textbf{G}$ and $\textbf{F}$, thus, rendering the required feature maps transductive:
\begin{align*}
\Tilde{\textbf{F}}^S = \textbf{G} \circ \textbf{F}^S \quad or \quad  \Tilde{\textbf{F}}^Q = \textbf{G} \circ \textbf{F}^Q.
\end{align*}
Here, $\textbf{F}^S$ and $\textbf{F}^Q$ represent the feature maps corresponding to the support and query images, respectively. As a result of operating on this channel-pixel distribution across images in a task, $\textbf{F}^S$ and $\textbf{F}^Q$ are rendered task-aware. Note that the query tensor $\textbf{Q}$ must not be confused with the query set $\mathcal{Q}$ of a task.
\subsection{Algorithmic Overview and Training Strategy}
\label{ssec:overview}
%
\begin{figure}[h]
\centering
\includegraphics[width=\columnwidth]{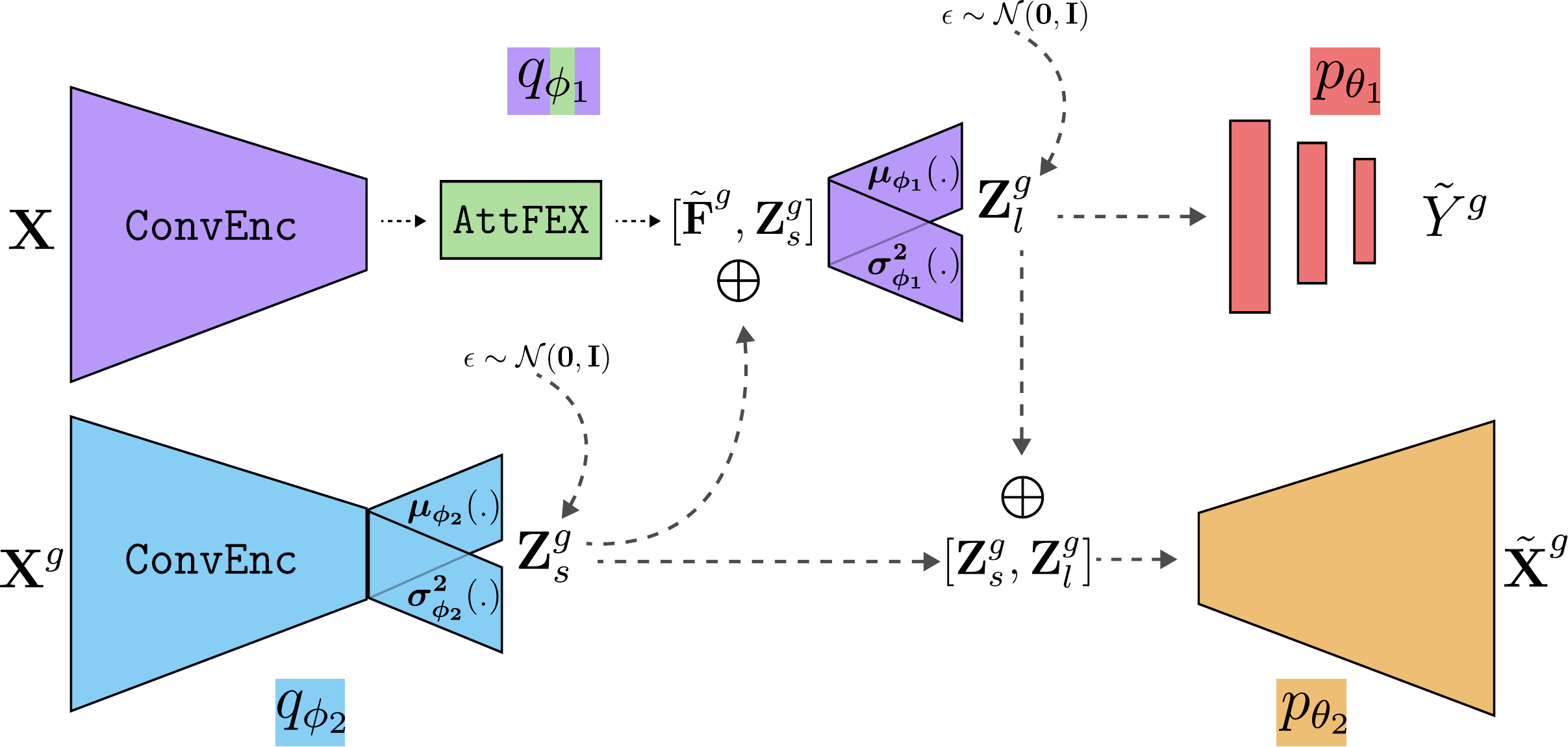}
\caption{\small \texttt{TRIDENT} is comprised of two intertwined variational networks. $\textbf{Z}_{s}^{g}$ is concatenated with the output of \texttt{AttFEX}, and used for inferring $\textbf{Z}_{l}^{g}$, where $g \in \{\mathcal{S}, \mathcal{Q}\}$. Next, both $\textbf{Z}_{l}^{g}$ and $\textbf{Z}_{s}^{g}$ are used to reconstruct images $\tilde{\textbf{X}}^g$ while $\textbf{Z}_{l}^{g}$ is used to extract $\tilde{Y}^{g}$.}
\label{fig:nn}
\end{figure}
\textbf{Overview of} \texttt{\textbf{TRIDENT}}. The complete architecture of \texttt{TRIDENT} is illustrated in Fig.~\ref{fig:nn}. The $\texttt{ConvEnc}$ feature extractor and the linear layers $\mu_{\phi_2}(.)$, $\sigma^2_{\phi_2}(.)$ constitute the inference network $q_{\phi_2}$ of the semantic latent variable (bottom row of Fig.~\ref{fig:nn}). The $\texttt{AttFEX}$ module, another $\texttt{ConvEnc}$, and linear layers $\mu_{\phi_1}(.)$ and $\sigma^2_{\phi_1}(.)$ make up the inference network $q_{\phi_1}$ of the label latent variable (top row of Fig.~\ref{fig:nn}). The proposed approach, \ourmethod, is described in Algorithm~\ref{alg:trident}. Note that \ourmethod{} is trained in a MAML \cite{pmlr-v70-finn17a} fashion, where depending on the inner or outer loop, the support or query set ($g \in \{\mathcal{S}, \mathcal{Q}\}$) will be the reference, respectively. First, the lower $\texttt{ConvEnc}$ block extracts feature maps  $\textbf{X}^{g}_{\texttt{CE}}=\texttt{ConvEnc}(\textbf{X}^{g})$. $\textbf{X}^{g}_{\texttt{CE}}$'s are then flattened and passed onto $\mu_{\phi_2}(.)$, $\sigma^2_{\phi_2}(.)$, which respectively output the mean and variance vectors of the \emph{semantic} latent distribution, as discussed in \eqref{eq:q1}. This is done either for the entire support or the query images $\textbf{X}^{g}$, where $g \in \{S, Q\}$ for a given task $\mathcal{T}_i$. We then sample a set of vectors $\textbf{Z}_{s}^{g}$ (subscript $s$ for \emph{semantic}) from their corresponding Gaussian distributions using the re-parameterization trick (line $1$, Algorithm~\ref{alg:trident}). Upon passing $\textbf{X} = \textbf{X}^{\mathcal{S}} \cup \textbf{X}^{\mathcal{Q}}$ through the upper \texttt{ConvEnc}, the \texttt{AttFEX} module of $q_{\phi_1}$ comes into play to create \emph{task-cognizant} feature maps $\tilde{\textbf{F}}^{g}$ for either $\mathcal{S}$ or $\mathcal{Q}$ (line $2$). $\textbf{Z}_{s}^{g}$ together with $\tilde{\textbf{F}}^g$ are passed onto the linear layers $\mu_{\phi_1}(.)$, $\sigma^2_{\phi_1}(.)$ to generate the mean and variance vectors of the \emph{label} latent Gaussian distributions (line $3$). After sampling the set of vectors $\textbf{Z}_{l}^{g}$ (subscript $l$ for \emph{label}) from their corresponding distributions, we use $\textbf{Z}_{l}^{g}$ and $\textbf{Z}_{s}^{g}$ to reconstruct the input images $\tilde{\textbf{X}}^{g}$ using the generative network $p_{\theta_2}$ (line 4). Next, $\textbf{Z}_{l}^{g}$'s are input to the classifier network $p_{\theta_1}$ to generate the class logits, which are normalized using a \texttt{softmax(.)}, resulting in class-conditional probabilities $p(\tilde{Y}^{g}\,\vert\,\textbf{Z}_{l}^{g})$ (line $5$). Finally (in line $6$), using the outputs of all the components discussed earlier, we calculate the loss $\mathcal{L}^g$ as formulated in \eqref{eq:loss} and \eqref{eq:nnloss}.
\begin{algorithm}[t!]
    \caption{\scalebox{.9}{\texttt{TRIDENT}}}\label{alg:trident}
    
    \setstretch{1.2}
    \SetKwInOut{Input}{input}
    \SetKwInOut{Output}{output}
    \SetKwInput{Require}{Require}
	\SetKwInput{Return}{Return}
	\SetKw{Let}{let}
	\SetKwRepeat{Do}{do}{while}
	
	\SetAlgoLined
	\LinesNumbered
	\DontPrintSemicolon
	\SetNoFillComment
	
    \Require{$\textbf{X}^{\mathcal{S}}, \textbf{X}^{\mathcal{Q}}, Y^g,$ $\textbf{X}^{g}_{\texttt{CE}},$ where $g \in \{\mathcal{S}, \mathcal{Q}\}$}
    
    Sample: \scalebox{.9}{$\textbf{Z}_{s}^{g} \sim q_{\phi_2}\big(\textbf{Z}_s \,\vert\, \boldsymbol{\mu}_{\phi_{2}}(\textbf{X}^{g}_{\texttt{CE}}),\, diag\big(\boldsymbol{\sigma}^{2}_{\phi_{2}}(\textbf{X}^{g}_{\texttt{CE}})\big)\big)$}\;
    Compute $\textit{task-cognizant}$ embeddings: \scalebox{.9}{$[ \Tilde{\textbf{F}}^\mathcal{S}, \Tilde{\textbf{F}}^\mathcal{Q}] = \texttt{AttFEX}(\texttt{ConvEnc}(\textbf{X})); \textbf{X} = \textbf{X}^{\mathcal{S}} \cup \textbf{X}^{\mathcal{Q}}$}\;
    Concatenate $\textbf{Z}_{s}^{g}$ and $\Tilde{\textbf{F}^{g}}$ into $[\Tilde{\textbf{F}^{g}}, \textbf{Z}_{s}^{g}]$ and sample: \scalebox{.9}{$\textbf{Z}_{l}^{g} \sim q_{\phi_1}\big(\textbf{Z}_{l} \,\vert\, \boldsymbol{\mu}_{\phi_{1}}([\Tilde{\textbf{F}}^g, \textbf{Z}_{s}^{g}]),\,  diag(\boldsymbol{\sigma}^{2}_{\phi_{1}}([\Tilde{\textbf{F}}^g, \textbf{Z}_{s}^{g}]))\big)$}\;
    Reconstruct \scalebox{.9}{$\textbf{X}^{g}$ using $\Tilde{\textbf{X}}^{g} = p_{\theta_2}(\textbf{X} \,\vert\, \textbf{Z}_{l}^{g}, \textbf{Z}_{s}^{g})$}\;
    Extract class-conditional probabilities using: \scalebox{.9}{$p\big(\tilde{Y}^{g} \,\vert\, \textbf{Z}_{l}^{g}\big) = \texttt{softmax}\big(p_{\theta_1}(Y^{g} \,\vert\, \textbf{Z}_{l}^{g})\big)$}\;
    Compute \scalebox{.9}{$\mathcal{L}^g = \mathcal{L}^g_R + \mathcal{L}^g_C$} using \eqref{eq:nnloss}\;
    \Return{$\mathcal{L}^g$}\vspace{+0.1cm}
\end{algorithm}
%
%
\newcounter{algorithm}
\setcounter{algorithm}{1}
\addtocounter{algorithm}{1}
\begin{algorithm}[t!]
    \caption{\scalebox{.9}{End to End Meta-Training of \texttt{TRIDENT}}}\label{alg:e2e-trident}
    
    \SetKwInOut{Input}{input}
    \SetKwInOut{Output}{output}
    \SetKwInput{Require}{Require}
	\SetKwInput{Return}{Return}
	\SetKw{Let}{let}
	\SetKwRepeat{Do}{do}{while}
	
	\SetAlgoLined
	\LinesNumbered
	\DontPrintSemicolon
	\SetNoFillComment
	
    \Require{$\mathcal{D}^{tr}$, $\alpha$, $\beta$, $B$}
    Randomly initialise $\Psi = (\phi_1, \phi_2, \theta_1, \theta_2)$\;
    
    \While {not converged}{
        Sample $B$ tasks $\mathcal{T}_i = \mathcal{S}_i \cup \mathcal{Q}_i$ from $\mathcal{D}^{tr}$\;
        \For{each task $\mathcal{T}_i$} {
            \For{number of adaptation steps}{
                Compute \scalebox{.9}{$\mathcal{L}^{\mathcal{S}_i}(\Psi) = \texttt{TRIDENT}(\mathcal{T}_i - \{Y^{\mathcal{Q}_i}\})$}\;
                Evaluate \scalebox{.9}{$\nabla_{(\Psi)}\mathcal{L}^{\mathcal{S}_i}(\Psi)$}\;
                \scalebox{.9}{$\Psi \gets \Psi - \alpha \nabla_{\Psi}\mathcal{L}^{\mathcal{S}_i}(\Psi)$}\;
            }
            \scalebox{.9}{$(\Psi')_{i} = \Psi$}
        }
        Compute \scalebox{.9}{$\mathcal{L}^{\mathcal{Q}_i}(\Psi'_{i}) = \texttt{TRIDENT}(\mathcal{T}_i - \{Y^{\mathcal{S}_i}\}); \forall i \in [1, B]$}\;
        Meta-update on $\mathcal{Q}_i$: \scalebox{.9}{$\Psi \gets \Psi - \beta \nabla_{\Psi} \sum_{i=1}^{B} \mathcal{L}^{\mathcal{Q}_i}(\Psi'_{i})$}\; 
    }
\end{algorithm}
%
%
\begin{table*}[t!]
\centering
\caption{\small Accuracies in (\% $\pm$ std). The predominant methodology of the baselines: \texttt{Ind.}: inductive inference, \texttt{TF}: transductive feature extraction methods, \texttt{TI}: transductive inference methods.  \texttt{Conv}: convolutional blocks, \texttt{RN}: \texttt{ResNet} backbone, $\dagger$: extra data. Style: \textbf{best} and \underline{second best}. \ourmethod{} employs a transductive feature extraction module (\texttt{TF}), and the simplest of backbones (\texttt{Conv4}).}
\vspace{-8pt}
\resizebox{\textwidth}{!}{%
\begin{tabular}{@{}lcccccccc@{}}
\toprule
                                            &  &                                          & \multicolumn{2}{|c|}{\cellcolor[HTML]{BEBDFF}\textbf{\textit{mini}Imagenet}}                                              & \multicolumn{2}{c|}{\cellcolor[HTML]{FAE0C1}\textbf{\textit{tiered}Imagenet}}
                                            & \multicolumn{2}{c}{\cellcolor[HTML]{C6C5C5}\textbf{\textit{mini}}$\rightarrow$\textbf{CUB}}\\ \midrule
\multicolumn{1}{l|}{\textbf{Methods}}       & \multicolumn{1}{c|}{\textbf{Backbone}} &
\multicolumn{1}{c|}{\textbf{Approach}} &
\multicolumn{1}{c|}{\textbf{5-way 1-shot}} & \multicolumn{1}{c|}{\textbf{5-way 5-shot}} & \multicolumn{1}{c|}{\textbf{5-way 1-shot}} & \multicolumn{1}{c|}{\textbf{5-way 5-shot}} & \multicolumn{1}{c|}{\textbf{5-way 1-shot}} & \textbf{5-way 5-shot} \\ \midrule
\multicolumn{1}{l|}{MAML \cite{pmlr-v70-finn17a}}               & \multicolumn{1}{c|}{$\texttt{Conv4}$} & \multicolumn{1}{c|}{$\texttt{Ind.}$}     & \multicolumn{1}{c|}{48.70 $\pm$ 1.84}         & \multicolumn{1}{c|}{63.11 $\pm$ 0.92}         & \multicolumn{1}{c|}{51.67 $\pm$ 1.81}         & \multicolumn{1}{c|}{70.30 $\pm$ 0.08}    &  \multicolumn{1}{c|}{34.01 $\pm$ 1.25}         & 48.83 $\pm$ 0.62    \\
\multicolumn{1}{l|}{ABML \cite{ravi2018amortized}}           & \multicolumn{1}{c|}{$\texttt{Conv4}$} & \multicolumn{1}{c|}{$\texttt{Ind.}$}   & \multicolumn{1}{c|}{40.88 $\pm$ 0.25}                 & \multicolumn{1}{c|}{58.19 $\pm$ 0.17}                 & \multicolumn{1}{c|}{-}                 & \multicolumn{1}{c|}{-}  & \multicolumn{1}{c|}{31.51 $\pm$ 0.32}                 & 47.80 $\pm$ 0.51\\
\multicolumn{1}{l|}{OVE(PL) \cite{DBLP:conf/nips/PatacchiolaTCOS20}}           & \multicolumn{1}{c|}{$\texttt{Conv4}$} & \multicolumn{1}{c|}{$\texttt{Ind.}$}   & \multicolumn{1}{c|}{48.00 $\pm$ 0.24}                 & \multicolumn{1}{c|}{67.14 $\pm$ 0.23}                 & \multicolumn{1}{c|}{-}                 & \multicolumn{1}{c|}{-}  & \multicolumn{1}{c|}{37.49 $\pm$ 0.11}                 & 57.23 $\pm$ 0.31\\
\multicolumn{1}{l|}{DKT+Cos \cite{DBLP:conf/nips/PatacchiolaTCOS20}}                    & \multicolumn{1}{c|}{$\texttt{Conv4}$} & \multicolumn{1}{c|}{$\texttt{Ind.}$}       & \multicolumn{1}{c|}{48.64 $\pm$ 0.45}           & \multicolumn{1}{c|}{62.85 $\pm$ 0.37} & \multicolumn{1}{c|}{-} & \multicolumn{1}{c|}{-}  & \multicolumn{1}{c|}{40.22 $\pm$ 0.54}                 & 55.65 $\pm$ 0.05\\
\multicolumn{1}{l|}{BOIL \cite{oh2021boil}}           & \multicolumn{1}{c|}{$\texttt{Conv4}$} & \multicolumn{1}{c|}{$\texttt{Ind.}$}   & \multicolumn{1}{c|}{49.61 $\pm$ 0.16}                 & \multicolumn{1}{c|}{48.58 $\pm$ 0.27}                 & \multicolumn{1}{c|}{66.45 $\pm$ 0.37}                 & \multicolumn{1}{c|}{69.37 $\pm$ 0.12} & \multicolumn{1}{c|}{-} & - \\
\multicolumn{1}{l|}{LFWT\cite{tseng2020cross}}      & \multicolumn{1}{c|}{\texttt{RN10}} & \multicolumn{1}{c|}{$\texttt{TF+TI}$} & \multicolumn{1}{c|}{66.32 $\pm$ 0.80}                 & \multicolumn{1}{c|}{81.98 $\pm$ 0.55}                  & \multicolumn{1}{c|}{-}                 & \multicolumn{1}{c|}{-} & \multicolumn{1}{c|}{47.47 $\pm$ 0.75}      & 66.98 $\pm$ 0.68     \\
\multicolumn{1}{l|}{FRN\cite{Wertheimer_2021_CVPR}}      & \multicolumn{1}{c|}{\texttt{RN12}} & \multicolumn{1}{c|}{$\texttt{Ind.}$} & \multicolumn{1}{c|}{66.45 $\pm$ 0.19}                 & \multicolumn{1}{c|}{82.83 $\pm$ 0.13}                  & \multicolumn{1}{c|}{71.16 $\pm$ 0.22}                 & \multicolumn{1}{c|}{86.01 $\pm$ 0.15} & \multicolumn{1}{c|}{54.11 $\pm$ 0.19}      & 77.09 $\pm$ 0.15      \\
\multicolumn{1}{l|}{DPGN\cite{Yang2020DPGNDP}}                   & \multicolumn{1}{c|}{$\texttt{RN12}$} & \multicolumn{1}{c|}{$\texttt{TF+TI}$}       & \multicolumn{1}{c|}{67.77}                 & \multicolumn{1}{c|}{84.6}                  & \multicolumn{1}{c|}{72.45}                 & \multicolumn{1}{c|}{87.24} & \multicolumn{1}{c|}{-} & -                 \\
\multicolumn{1}{l|}{PAL\cite{ma2021partner}}           & \multicolumn{1}{c|}{$\texttt{RN12}$} & \multicolumn{1}{c|}{$\texttt{TF+TI}$}   & \multicolumn{1}{c|}{69.37 $\pm$ 0.64}                 & \multicolumn{1}{c|}{84.40 $\pm$ 0.44}                 & \multicolumn{1}{c|}{72.25 $\pm$ 0.72}                 & \multicolumn{1}{c|}{86.95 $\pm$ 0.47} & \multicolumn{1}{c|}{-} & - \\
\multicolumn{1}{l|}{Proto-Completion\cite{zhang2021prototype}}           & \multicolumn{1}{c|}{$\texttt{RN12}$} & \multicolumn{1}{c|}{$\texttt{TF+TI}$}   & \multicolumn{1}{c|}{73.13 $\pm$ 0.85}           & \multicolumn{1}{c|}{82.06 $\pm$ 0.54}           & \multicolumn{1}{c|}{81.04 $\pm$ 0.89}           & \multicolumn{1}{c|}{87.42 $\pm$ 0.57} & \multicolumn{1}{c|}{-} & -           \\
\multicolumn{1}{l|}{TPMN\cite{wu2021task}}           & \multicolumn{1}{c|}{$\texttt{RN12}$} & \multicolumn{1}{c|}{$\texttt{TF+TI}$}   & \multicolumn{1}{c|}{67.64 $\pm$ 0.63}                 & \multicolumn{1}{c|}{83.44 $\pm$ 0.43}                 & \multicolumn{1}{c|}{72.24 $\pm$ 0.70}                 & \multicolumn{1}{c|}{86.55 $\pm$ 0.63} & \multicolumn{1}{c|}{-} & - \\
\multicolumn{1}{l|}{LIF-EMD\cite{Li_Wang_Hu_2021}}           & \multicolumn{1}{c|}{$\texttt{RN12}$} & \multicolumn{1}{c|}{$\texttt{TF+TI}$}   & \multicolumn{1}{c|}{68.94 $\pm$ 0.28}                 & \multicolumn{1}{c|}{85.07 $\pm$ 0.50}                 & \multicolumn{1}{c|}{73.76 $\pm$ 0.32}                 & \multicolumn{1}{c|}{87.83 $\pm$ 0.59} & \multicolumn{1}{c|}{-} & - \\
\multicolumn{1}{l|}{Transd-CNAPS\cite{Bateni2022_TransductiveCNAPS}}           & \multicolumn{1}{c|}{$\texttt{RN18}$} & \multicolumn{1}{c|}{$\texttt{TF+TI}$}   & \multicolumn{1}{c|}{55.6 $\pm$ 0.9}           & \multicolumn{1}{c|}{73.1 $\pm$ 0.7}           & \multicolumn{1}{c|}{65.9 $\pm$ 1.0}           & \multicolumn{1}{c|}{81.8 $\pm$ 0.7} & \multicolumn{1}{c|}{-} & -           \\

\multicolumn{1}{l|}{Baseline++\cite{chen19closerfewshot}}                    & \multicolumn{1}{c|}{$\texttt{RN18}$}  & \multicolumn{1}{c|}{$\texttt{TF}$}  & \multicolumn{1}{c|}{51.87 $\pm$ 0.77}                  & \multicolumn{1}{c|}{75.68 $\pm$ 0.63}                 & \multicolumn{1}{c|}{-}                 & \multicolumn{1}{c|}{-} & \multicolumn{1}{c|}{42.85 $\pm$ 0.69} & 62.04 $\pm$ 0.76                  \\
\multicolumn{1}{l|}{FEAT\cite{ye2020fewshot}}           & \multicolumn{1}{c|}{$\texttt{RN18}$} & \multicolumn{1}{c|}{$\texttt{TF}$}   & \multicolumn{1}{c|}{66.78}                 & \multicolumn{1}{c|}{82.05}                 & \multicolumn{1}{c|}{70.80}                 & \multicolumn{1}{c|}{84.79} & \multicolumn{1}{c|}{50.67 $\pm$ 0.78} &  71.08 $\pm$ 0.73                 \\
\multicolumn{1}{l|}{SimpleShot\cite{wang2019simpleshot}}             & \multicolumn{1}{c|}{$\texttt{WRN}$}  & \multicolumn{1}{c|}{$\texttt{Ind.}$}      & \multicolumn{1}{c|}{63.32}                 & \multicolumn{1}{c|}{80.28}                 & \multicolumn{1}{c|}{69.98}                 & \multicolumn{1}{c|}{85.45} & \multicolumn{1}{c|}{48.56 } & 65.63                 \\
\multicolumn{1}{l|}{Assoc-Align\cite{afrasiyabi2020associative}}           & \multicolumn{1}{c|}{$\texttt{WRN}$} & \multicolumn{1}{c|}{$\texttt{TF}$}   & \multicolumn{1}{c|}{65.92 $\pm$ 0.60}                 & \multicolumn{1}{c|}{82.85 $\pm$ 0.55}                 & \multicolumn{1}{c|}{74.40 $\pm$ 0.68}                 & \multicolumn{1}{c|}{86.61 $\pm$ 0.59} & \multicolumn{1}{c|}{47.25 $\pm$ 0.76}                  & 72.37 $\pm$ 0.89\\
\multicolumn{1}{l|}{ReRank\cite{shen2021reranking}}           & \multicolumn{1}{c|}{$\texttt{WRN}$} & \multicolumn{1}{c|}{$\texttt{TF+TI}$}   & \multicolumn{1}{c|}{72.4$\pm$0.6}                 & \multicolumn{1}{c|}{80.2$\pm$0.4}                 & \multicolumn{1}{c|}{79.5$\pm$0.6}                 & \multicolumn{1}{c|}{84.8$\pm$0.4} & \multicolumn{1}{c|}{-}                 & -  \\
\multicolumn{1}{l|}{TIM-GD\cite{NEURIPS2020_196f5641}}                    & \multicolumn{1}{c|}{$\texttt{WRN}$} & \multicolumn{1}{c|}{$\texttt{TI}$}   & \multicolumn{1}{c|}{77.8}                  & \multicolumn{1}{c|}{87.4}                 & \multicolumn{1}{c|}{82.1}                 & \multicolumn{1}{c|}{89.8} & \multicolumn{1}{c|}{-} & 71                  \\
\multicolumn{1}{l|}{LaplacianShot\cite{DBLP:conf/icml/ZikoDGA20}}          & \multicolumn{1}{c|}{$\texttt{WRN}$} & \multicolumn{1}{c|}{$\texttt{TI}$}    & \multicolumn{1}{c|}{74.9}                  & \multicolumn{1}{c|}{84.07}                 & \multicolumn{1}{c|}{80.22}                 & \multicolumn{1}{c|}{87.49} & \multicolumn{1}{c|}{55.46}                 & 66.33                 \\
\multicolumn{1}{l|}{S2M2\cite{mangla2020charting}}          & \multicolumn{1}{c|}{$\texttt{WRN}$} & \multicolumn{1}{c|}{$\texttt{TF}$}   & \multicolumn{1}{c|}{64.93 $\pm$ 0.18}                  & \multicolumn{1}{c|}{83.18 $\pm$ 0.11}                 & \multicolumn{1}{c|}{73.71 $\pm$ 0.22}                 & \multicolumn{1}{c|}{88.59 $\pm$ 0.14} & \multicolumn{1}{c|}{48.24 $\pm$ 0.84}                 & 70.44 $\pm$ 0.75                 \\
\multicolumn{1}{l|}{MetaQDA\cite{zhang2021shallow}}          & \multicolumn{1}{c|}{$\texttt{WRN}$} & \multicolumn{1}{c|}{$\texttt{TF}$}   & \multicolumn{1}{c|}{67.83 $\pm$ 0.64}                  & \multicolumn{1}{c|}{ 84.28 $\pm$ 0.69}                 & \multicolumn{1}{c|}{74.33 $\pm$ 0.65}                 & \multicolumn{1}{c|}{89.56 $\pm$ 0.79} & \multicolumn{1}{c|}{53.75 $\pm$ 0.72}                 & 71.84 $\pm$ 0.66                 \\
\multicolumn{1}{l|}{PT+MAP\cite{hu2021leveraging}}          & \multicolumn{1}{c|}{$\texttt{WRN}$} & \multicolumn{1}{c|}{$\texttt{TF+TI}$}   & \multicolumn{1}{c|}{82.92 $\pm$ 0.26}                  & \multicolumn{1}{c|}{ 88.82  $\pm$ 0.13}                 & \multicolumn{1}{c|}{85.67 $\pm$ 0.26}                 & \multicolumn{1}{c|}{90.45 $\pm$ 0.14} & \multicolumn{1}{c|}{62.49 $\pm$ 0.32}                 & 76.51 $\pm$ 0.18                 \\
\multicolumn{1}{l|}{PE$M_n$E-BMS\cite{hu2022squeezing}}          & \multicolumn{1}{c|}{$\texttt{WRN}$} & \multicolumn{1}{c|}{$\texttt{TF+TI}$}   & \multicolumn{1}{c|}{\underline{83.35 $\pm$ 0.25}}                  & \multicolumn{1}{c|}{ 89.53 $\pm$ 0.13}                 & \multicolumn{1}{c|}{\underline{86.07 $\pm$ 0.25}}                 & \multicolumn{1}{c|}{\underline{91.09 $\pm$ 0.14}} & \multicolumn{1}{c|}{\underline{63.90 $\pm$ 0.31}}                 & \underline{79.15 $\pm$ 0.18}                 \\
\multicolumn{1}{l|}{Transd-CNAPS+FETI\cite{Bateni2022_TransductiveCNAPS}}          & \multicolumn{1}{c|}{$\texttt{RN18}^\dagger$} & \multicolumn{1}{c|}{$\texttt{TF+TI}$}   & \multicolumn{1}{c|}{79.9 $\pm$ 0.8}                  & \multicolumn{1}{c|}{ \underline{91.50 $\pm$ 0.4}}                 & \multicolumn{1}{c|}{73.8 $\pm$ 0.1}                 & \multicolumn{1}{c|}{87.7 $\pm$ 0.6} & \multicolumn{1}{c|}{-}                 & -                 \\
\midrule
\rowcolor{LightCyan} \multicolumn{1}{l|}{\textbf{\texttt{TRIDENT}\textbf{(Ours)}}}  & \multicolumn{1}{c|}{$\texttt{Conv4}$} & \multicolumn{1}{c|}{$\texttt{TF}$}      & \multicolumn{1}{c|}{\textbf{86.11 $\pm$ 0.59}}         & \multicolumn{1}{c|}{\textbf{95.95 $\pm$ 0.28}}         & \multicolumn{1}{c|}{\textbf{86.97 $\pm$ 0.50}}          & \multicolumn{1}{c|}{\textbf{96.57 $\pm$ 0.17}}    & \multicolumn{1}{c|}{\textbf{84.61 $\pm$ 0.33}} & \textbf{80.74 $\pm$ 0.35}\\ \bottomrule
\end{tabular}%
}
\label{tab:mini-tier-cross}
\end{table*}
\noindent\textbf{Training strategy.} An important aspect of the training procedure of \texttt{TRIDENT} is that its set of parameters $\Psi = (\theta_1, \theta_2, \phi_1, \phi_2)$ are meta-learnt by back-propagating through the adaptation procedure on the support set, as proposed in MAML \cite{pmlr-v70-finn17a} and illustrated here in Algorithm~\ref{alg:e2e-trident}. This increases the sensitivity of the parameters $\Psi$ towards the loss function for fast adaptation to unseen tasks and reduces generalization errors on the query set $\mathcal{Q}$.  
First, we randomly initialize the parameters $\Psi$ (line 1, Algorithm~\ref{alg:e2e-trident}) to compute the objective function over the support set $\mathcal{L}^{\mathcal{S}_i}(\Psi)$ using equation (3) in the main manuscript, and perform a number of gradient descent steps on the parameters $\Psi$ to adapt them to the support set (lines $5$ to $9$). This is called the \emph{inner-update} and is done separately for all the support sets corresponding to their $B$ different tasks (line $3$). Once the inner-update is computed for each of the $B$ parameter sets, the loss is evaluated on the query set $\mathcal{L}^{\mathcal{Q}_i}(\Psi'_{i})$ (line $12$), following which a \emph{meta-update} is conducted over all the corresponding query sets, which involves computing a gradient through a gradient procedure as described in \cite{pmlr-v70-finn17a} (line $13$).
\section{Experimental Evaluation}
The goal of this section is to address the following four questions: (i) How well does $\texttt{TRIDENT}$ perform when compared against the state-of-the-art methods for few-shot classification? (ii) How reliable is $\texttt{TRIDENT}$ in terms of the confidence and uncertainty metrics? (iii) How well does $\texttt{TRIDENT}$ perform in a cross-domain setting where there is a domain shift between the training and testing datasets? (iv) Does \ourmethod{} actually decouple latent variables? 

\noindent \textbf{Benchmark Datasets.} We evaluate \ourmethod{} on the three most commonly adopted datasets: \textit{mini}Imagenet \cite{Ravi2017OptimizationAA}, \textit{tiered}Imagenet \cite{ren2018metalearning} and CUB \cite{WelinderEtal2010}. \textit{mini}Imagenet and \textit{tiered}Imagenet are subsets of ImageNet \cite{deng2009imagenet} utilized for few-shot classification. Further details on these datasets can be found in the Appendix.

\noindent \textbf{Implementational Details.} We use PyTorch \cite{NEURIPS2019_9015} and learn2learn \cite{Arnold2020-ss} for all our implementations. We use a commonly adopted \texttt{Conv4} architecture \cite{Ravi2017OptimizationAA, pmlr-v70-finn17a, DBLP:conf/nips/PatacchiolaTCOS20, afrasiyabi2020associative, wang2019simpleshot, NEURIPS2020_196f5641} as \texttt{ConvEnc} to obtain the generic feature maps. Following the standard setting in the literature \cite{pmlr-v70-finn17a, Ravi2017OptimizationAA}, the \texttt{Conv4} has four convolutional blocks where each block has a $3 \times 3$ convolution layer with 32 feature maps, followed by a batch normalization (BN) \cite{pmlr-v37-ioffe15} layer, a $2 \times 2$ max-pooling layer and a \texttt{LeakyReLU(0.2)} activation. The generative network $p_{\theta_1}$ for $\textbf{z}_l$ is a classifier with two linear layers and a \texttt{LeakyReLU(0.2)} activation in between, while $p_{\theta_2}$ for $\textbf{z}_s$ consists of four blocks of a $2$-D upsampling layer, followed by a $3 \times 3$ convolution and \texttt{LeakyReLU(0.2)} activation. Both latent variables $\textbf{z}_l$ and $\textbf{z}_s$ have a dimensionality of $64$. Following \cite{DBLP:journals/corr/abs-1803-02999, liu2019fewTPN, NIPS2017_3f5ee243}, images are resized to $84\times84$ for all configurations and we train and report test accuracy of ($5$-way, $1$ and $5$-shot) settings with $10$ query images per class for all datasets. Hyper-parameter settings can be found in the Appendix.  
\subsection{Evaluation Results}
We report test accuracies indicating $95$\% confidence intervals over $600$ tasks for \textit{mini}Imagenet, and $2000$ tasks for both \textit{tiered}Imagenet and CUB, as is customary across the literature \cite{chen19closerfewshot, Dhillon2020A, Bateni2022_TransductiveCNAPS}. We compare our performance against a wide variety of state-of-the-art few-shot classification methods such as: (i) metric-learning \cite{wang2019simpleshot, Bateni2020_SimpleCNAPS, afrasiyabi2020associative, Yang2020DPGNDP}, (ii) transductive feature-extraction based \cite{NEURIPS2018_66808e32, ye2020fewshot, li2019ctm, xu2021attentional}, (iii) optimization-based \cite{pmlr-v70-finn17a, mishra2018a, oh2021boil, lee2019meta, rusu2018metalearning}, (iv) transductive inference-based \cite{Bateni2022_TransductiveCNAPS, NEURIPS2020_196f5641, DBLP:conf/icml/ZikoDGA20, liu2019fewTPN}, and (v) Bayesian \cite{pmlr-v119-iakovleva20a, 9010263, Hu2020Empirical, DBLP:conf/nips/PatacchiolaTCOS20, ravi2018amortized} approaches. Previous works \cite{liu2019fewTPN}, \cite{NEURIPS2019_01894d6f} have demonstrated the superiority of transductive inference methods over their inductive counterparts. In this light, we compare against a larger number of transductive ($18$ baselines) rather than inductive ($7$ baselines) methods for a fair comparison. 
\begin{table}[t!]
\centering
\vspace{-0.5cm}
\caption{\small Parameter count of \texttt{TRIDENT} against competitors.}
\vspace{-8pt}
\label{params}
\resizebox{\columnwidth}{!}{%
\begin{tabular}{c|cccc|>{\columncolor[HTML]{D8FEFD}}c|c|c|c}
\hline
 &
  \multicolumn{1}{c|}{\texttt{Conv4}} &
  \multicolumn{1}{c|}{$\mu_{\phi}$} &
  \multicolumn{1}{c|}{$\sigma_{\phi}$} &
  \texttt{AttFEX} &
  \texttt{\textbf{TRIDENT}} &
  \texttt{Conv4} &
  \multicolumn{1}{c|}{\texttt{RN18}} &
   \texttt{WRN} \\ \hline
$q_{\phi_1}$ & \multicolumn{1}{c|}{28896} & \multicolumn{1}{c|}{51264} & \multicolumn{1}{c|}{51264} & 6994 & \cellcolor[HTML]{D8FEFD} &  & \\ \cline{1-5}
$q_{\phi_2}$ & \multicolumn{1}{c|}{28896} & \multicolumn{1}{c|}{51264} & \multicolumn{1}{c|}{51264} & -    & \cellcolor[HTML]{D8FEFD} &  & \\ \cline{1-5}
$p_{\theta_1}$+ $p_{\theta_2}$ &
  \multicolumn{4}{c|}{2245 + 132009} &
  \multirow{-3}{*}{\cellcolor[HTML]{D8FEFD}$\textbf{412,238}$} &
  \multirow{-3}{*}{$190,410$} &
  \multirow{-3}{*}{$12.4$M} &
  \multirow{-3}{*}{$36.482$M}\\ \hline
\end{tabular}%
}
\vspace{-0.5cm}
\end{table}

It is important to note that \ourmethod{} is only a \emph{transductive feature-extraction} based method as we utilize the query set images to extract task-aware feature embeddings; it is not a transductive inference based method since we perform inference of class-labels over the entire domain of definition and not just for the selected query samples \cite{Vapnik2006EstimationOD, gammerman1998learning}. The results on \textit{mini}Imagenet and \textit{tiered}Imagenet for both ($5$-way, $1$ and $5$-shot) settings are summarized in Table~\ref{tab:mini-tier-cross}. We accentuate on the fact that we also compare against Transd-CNAPS+FETI \cite{Bateni2022_TransductiveCNAPS}, where the authors pre-train the \texttt{ResNet-18} backbone on the entire train split of Imagenet. We, however, avoid training on additional datasets, in favor of fair comparison with the rest of literature. Regardless of the choice of backbone (simplest in our case), \texttt{TRIDENT} sets a new state-of-the-art on \textit{mini}Imagenet and \textit{tiered}Imagenet for both ($5$-way, $1$ and $5$-shot) settings, offering up to $5\%$ gain over the prior art. Recently, a more challenging \emph{cross-domain} setting has been proposed for few-shot classification to assess its generalization capabilities to unseen datasets. The commonly adopted setting is where one trains on \textit{mini}Imagenet and tests on CUB \cite{chen19closerfewshot}. The results of this experiment are also presented in Table~\ref{tab:mini-tier-cross}. We compare against \emph{any existing baselines} for which this cross-domain experiment has been conducted. As can be seen, and to the best of our knowledge, \texttt{TRIDENT} again sets a new state-of-the-art by a significant margin of $20\%$ for ($5$-way, $1$-shot) setting, and $1.5\%$ for ($5$-way, $5$-shot) setting.

\noindent \textbf{Computational Complexity.}  Most of the reported baselines in Table~\ref{tab:mini-tier-cross} use stronger backbones such as \texttt{ResNet12}, \texttt{ResNet18} and \texttt{WRN} which contain $11.5$, $12.4$ and $36.4$ millions of parameters respectively. On the other hand, we use three \texttt{Conv4}s along with two fully connected layers and an \texttt{AttFEX} module which accounts for 410,958 and 412,238 parameters in the ($5$-way, $1$-shot) and ($5$-way, $5$-shot) scenarios, respectively. This is summarized in details in Table~\ref{params}. Even though we are more parameter heavy than approaches that use a single \texttt{Conv4} as feature extractor, \ourmethod's total parameters still lies in the same order of magnitude as these approaches. In summary, when it comes to complexity in parameter space, we are considerably more efficient than the vast majority of the cited competitors. 

\noindent \textbf{Reliability Metrics.} A complementary set of metrics are typically used in  probabilistic settings to measure the uncertainty and reliability of predictions. More specifically, expected calibration error (ECE) and maximum calibration error (MCE) respectively measure the expected and maximum binned difference between confidence and accuracy \cite{pmlr-v70-guo17a}. This is illustrated in Table~\ref{tab:rel} where \ourmethod{} offers superior calibration on \textit{mini}Imagenet ($5$-way, $1$ and $5$-shot) as compared to other probabilistic approaches, and MAML \cite{pmlr-v70-finn17a}. 
\subsection{Ablation Study}
We analyze the classification performance of \ourmethod{} across various paramaters and hyper-parameters, as is summarized in Table~\ref{tab:ablation}. We use \textit{mini}Imagenet ($5$-way, $1$-shot) setting to carry out ablation study experiments. To cover different design perspectives, we carry out ablation on: (i) MAML-style training parameters: meta-batch size $B$ and number of inner adaption steps $n$, (ii) latent space dimensionality: $\textbf{z}_l$ and $\textbf{z}_s$ to assess the impact of their size, (iii) \ourmodule{} features: number of features extracted by $\textbf{W}_M$, $\textbf{W}_N$. Looking at the results,  \ourmethod{}'s performance is directly proportional to the number of tasks and inner-adaptation steps, as is previously demonstrated in \cite{AntoniouES19, pmlr-v70-finn17a} for MAML based training. Regarding latent space dimensions, a correlation between a higher dimension of $\textbf{z}_l$ and $\textbf{z}_s$ and a better performance can be observed. Even though, the results show that increasing both dimensions beyond $64$ leads to performance degradation. As such, $(64, 64)$ seems to be the sweet spot. Finally, on feature space dimensions of \ourmodule, the performance improves when $\textbf{W}_M$ $>$ $\textbf{W}_N$, and the best performance is achieved when the parameters are set to ($64$, $32$). Notably, the exact set of parameters return the best performance for ($5$-way, $5$-shot) setting.
\begin{table}[t!]
\vspace{-0.5cm}
\centering
\caption{\small Calibration errors of \ourmethod{}. Style: \textbf{best} and \underline{second best}.}
\vspace{-8pt}
\label{tab:rel}
\resizebox{\columnwidth}{!}{%
\begin{tabular}{@{}l|c|c|c|c|c|c|c|
>{\columncolor[HTML]{D8FEFD}}c| @{}}
\toprule
                      & \textbf{Metrics} & MAML  & PLATIPUS & ABPML             & ABML  & BMAML & VAMPIRE           & \textbf{\texttt{TRIDENT}} \\ \midrule
\multicolumn{1}{l|}{} & \textbf{ECE}     & 0.046 & 0.032    & 0.013             & 0.026 & 0.025 & \underline{0.008} & \textbf{0.0036}           \\ \cmidrule(l){2-9} 
\multicolumn{1}{l|}{\multirow{-2}{*}{\begin{tabular}[c]{@{}l@{}}\textbf{5-way},\\ \textbf{1-shot}\end{tabular}}} &
  \textbf{MCE} &
  0.073 &
  0.108 &
  \underline{0.037} &
  0.058 &
  0.092 &
  0.038 &
  \textbf{0.029} \\ \midrule
\multicolumn{1}{l|}{} & \textbf{ECE}     & 0.032 & -        & \underline{0.006} & -     & 0.027 & -                 & \textbf{0.0015}           \\ \cmidrule(l){2-9} 
\multicolumn{1}{l|}{\multirow{-2}{*}{\begin{tabular}[c]{@{}l@{}}\textbf{5-way},\\ \textbf{5-shot}\end{tabular}}} &
  \textbf{MCE} &
  0.044 &
  - &
  \underline{0.030} &
  - &
  0.049 &
  - &
  \textbf{0.018} \\ \bottomrule
\end{tabular}%
}
\vspace{-0.5cm}
\end{table}
\begin{table*}[t!]
\centering
\caption{\small Ablation study for \textit{mini}Imagenet ($5$-way, $1$-shot) tasks. Accuracies in (\% $\pm$ std.).}
\vspace{-8pt}
\label{tab:ablation}
\resizebox{\textwidth}{!}{%
\begin{tabular}{@{}l|c|cc|c|cc|c|cc@{}}
\toprule
\multirow{2}{*}{\textbf{(B, n)}} &
  \textbf{(5, 3)} &
  \multicolumn{2}{c|}{\textbf{(5, 5)}} &
  \textbf{(10, 3)} &
  \multicolumn{2}{c|}{\textbf{(10, 5)}} &
  \textbf{(20, 3)} &
  \multicolumn{2}{c}{\textbf{(20, 5)}} \\ \cmidrule(l){2-10} 
 &
  - &
  \multicolumn{2}{c|}{67.43 $\pm$ 0.75} &
  69.21 $\pm$ 0.66 &
  \multicolumn{2}{c|}{74.6 $\pm$ 0.84} &
  80.82 $\pm$ 0.68 &
  \multicolumn{2}{c}{\textbf{86.11 $\pm$ 0.59}} \\ \midrule \midrule
\multirow{2}{*}{\begin{tabular}[c]{@{}l@{}}($dim(\textbf{z}_l)$, \\  $dim(\textbf{z}_s))$\end{tabular}} &
  \textbf{(32, 32)} &
  \multicolumn{1}{c|}{\textbf{(32, 64)}} &
  \textbf{(32, 128)} &
  \textbf{(64, 32)} &
  \multicolumn{1}{c|}{\textbf{(64, 64)}} &
  \textbf{(64, 128)} &
  \textbf{(128, 32)} &
  \multicolumn{1}{c|}{\textbf{(128, 64)}} &
  \textbf{(128, 128)} \\ \cmidrule(l){2-10} 
 &
  76.29 $\pm$ 0.72 &
  \multicolumn{1}{c|}{75.44 $\pm$ 0.81} &
  79.1 $\pm$ 0.57 &
  82.93 $\pm$ 0.8 &
  \multicolumn{1}{c|}{\textbf{86.11 $\pm$ 0.59}} &
  85.62 $\pm$ 0.52 &
  81.49 $\pm$ 0.65 &
  \multicolumn{1}{c|}{82.89 $\pm$ 0.48} &
  84.42 $\pm$ 0.59 \\ \midrule \midrule
\multirow{2}{*}{\begin{tabular}[c]{@{}l@{}}($dim(\textbf{W}_M)$, \\  $dim(\textbf{W}_N))$\end{tabular}} &
  \textbf{(32, 32)} &
  \multicolumn{1}{c|}{\textbf{(32, 64)}} &
  \textbf{(32, 128)} &
  \textbf{(64, 32)} &
  \multicolumn{1}{c|}{\textbf{(64, 64)}} &
  \textbf{(64, 128)} &
  \textbf{(128, 32)} &
  \multicolumn{1}{c|}{\textbf{(128, 64)}} &
  \textbf{(128, 128)} \\ \cmidrule(l){2-10} 
 &
  78.4 $\pm$ 0.23 &
  \multicolumn{1}{c|}{77.89 $\pm$ 0.39} &
  79.55 $\pm$ 0.87 &
  \textbf{86.11 $\pm$ 0.59} &
  \multicolumn{1}{c|}{84.87 $\pm$ 0.45} &
  82.11 $\pm$ 0.35 &
  84.67 $\pm$ 0.7 &
  \multicolumn{1}{c|}{85.8 $\pm$ 0.58} &
  83.92 $\pm$ 0.63 \\ \bottomrule
\end{tabular}%
}
\end{table*}
%
\subsection{Decoupling Analysis}
\begin{figure}[t!]
\centering
\includegraphics[width=\columnwidth]{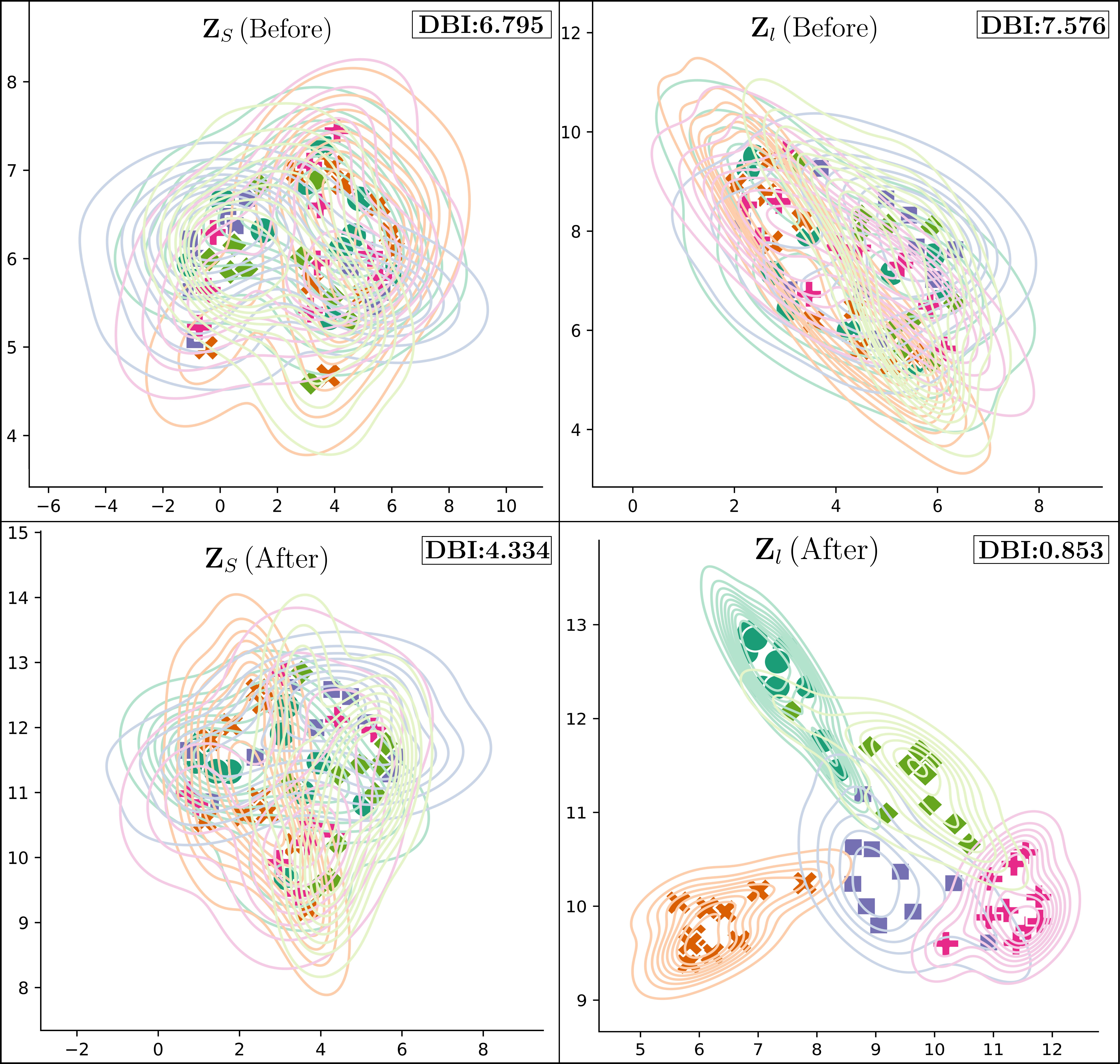}
\caption{\small Better class separation upon meta-update is confirmed by lower DBI scores. Different colors/markers indicate classes.}
\label{fig:latents}
\end{figure}

As a qualitative demonstration, we visualize the \emph{label} and \emph{semantic} latent means ($\mu_l$ and $\mu_s$) of query images for a randomly selected ($5$-way, $5$-shot) task from \textit{mini}Imagenet, before and after the MAML meta-update procedure. The \texttt{UMAP} \cite{mcinnes2018umap} plots in Fig.~\ref{fig:latents} illustrate significant improvement in class-conditional separation of query samples for \emph{label} latent space upon meta-adaption, whereas negligible improvement is visible on the semantic latent space. This is a qualitative evidence that $\textbf{Z}_L$ captures more class-discriminating information as compared to $\textbf{Z}_S$. To substantiate this quantitatively, the clustering capacity of these latent spaces is also measured by the Davies-Bouldin score (DBI) \cite{4766909}, where, the lower the DBI score, the better both the inter-cluster separation and intra-cluster ``tightness". Fig.~\ref{fig:latents} shows that the DBI score drops significantly more after meta-adaptation in the case of $\textbf{Z}_L$ as compared to $\textbf{Z}_S$, indicating better clustering of features in the former than the latter. This aligns with the proposed decoupling strategy of \texttt{TRIDENT} and corroborates the validity of our proposition to put an emphasis on label latent information for the downstream few-shot task. 
%
\section{Concluding Remarks}
We introduce a novel variational inference network (coined as \ourmethod) that simultaneously infers decoupled latent variables representing semantic and label information of an image. The proposed network is comprised of two intertwined variational sub-networks responsible for inferring the semantic and label information separately, the latter being enhanced using an attention-based transductive feature extraction module (\ourmodule). Our extensive experimental results corroborate the efficacy of this transductive decoupling strategy on a variety of few-shot classification settings demonstrating superior performance and setting a new state-of-the-art for the most commonly adopted dataset \textit{mini} and \textit{tiered}Imagenet as well as for the recent challenging cross-domain scenario of \textit{mini}Imagenet $\rightarrow$ CUB. As future work, we plan to demonstrate the applicability of \ourmethod{} in semi-supervised and unsupervised settings by including the likelihood of unlabelled samples derived from the graphical model. This would render \texttt{TRIDENT} as an all-inclusive holistic approach towards solving few-shot classification. 

\begin{quote}
\begin{small}
\bibliography{TRIDENT}
\end{small}
\end{quote}

\clearpage
\appendix
\section{Appendix}
\subsection{Impact of \textbf{\ourmodule}}
In order to study the impact of the transductive feature extractor \ourmodule{}, we exclude it during training and train the remaining architecture. Training proceeds exactly as mentioned before in the manuscript.
\begin{table}[h]
\centering
\caption{\small Impact of \ourmodule{} on classification accuracies.}
\vspace{-0.2cm}
\label{tab:wo attfex}
\resizebox{\columnwidth}{!}{%
\begin{tabular}{@{}ccccc@{}}
\toprule
 &
  \multicolumn{2}{|c|}{\cellcolor[HTML]{BEBDFF}\textbf{\textit{mini}Imagenet}} &
  \multicolumn{2}{c}{\cellcolor[HTML]{FAE0C1}\textbf{\textit{tiered}Imagenet}} \\ \midrule
\multicolumn{1}{l|}{} &
  \multicolumn{1}{c|}{\textbf{(5-way, 1-shot)}} &
  \multicolumn{1}{c|}{\textbf{(5-way, 5-shot)}} &
  \multicolumn{1}{c|}{\textbf{(5-way, 1-shot)}} &
  \textbf{(5-way, 5-shot)} \\ \midrule
\multicolumn{1}{c|}{\textbf{\ourmodule{} OFF}} &
  \multicolumn{1}{c|}{67.68 $\pm$ 0.55} &
  \multicolumn{1}{c|}{78.53 $\pm$ 0.21} &
  \multicolumn{1}{c|}{69.32 $\pm$ 0.76} &
  79.32 $\pm$ 0.76 \\ \midrule
\multicolumn{1}{c|}{\textbf{\ourmodule{} ON}} &
  \multicolumn{1}{c|}{86.11 $\pm$ 0.59} &
  \multicolumn{1}{c|}{95.95 $\pm$ 0.28} &
  \multicolumn{1}{c|}{86.97 $\pm$ 0.50} &
  96.57 $\pm$ 0.17 \\ \bottomrule
\end{tabular}%
}
\end{table}
As can be seen in Table~\ref{tab:wo attfex}, the exclusion of \ourmodule{} from \ourmethod{} results in a substantial drop in classification performance across both datasets and task settings. Empirically, this further substantiates the importance of \ourmodule{}'s ability to render the feature maps transductive/task-aware. As explained earlier in the main manuscript, it is imperative to include $y$ in the input to $q_{\phi_1}(.)$ for mathematical correctness of the variational inference formulation. However, in order to utilize \ourmethod{} as a classification and not a label reconstruction network, we choose not to input $y$ to $q_{\phi_1}(.)$, but rather do so indirectly by inducing a semblance of label characteristics in the features extracted from the images in a task. Thus, it is important to realize that this ability of \ourmodule{} to render feature maps transductive is not just an adhoc performance enhancer, but rather an essential part of \ourmethod{} since it allows us to not violate our generative and inference mechanics. 
\subsection{Additional Details of Datasets}
\textbf{\textit{mini}Imagenet} \cite{NIPS2016_90e13578} is a subset of ImageNet \cite{deng2009imagenet} for few-shot classification. It contains $100$ classes with $600$ samples each. We follow the predominantly adopted settings of \cite{Ravi2017OptimizationAA, chen19closerfewshot} where we split the entire dataset into $64$ classes for training, $16$ for validation and $20$ for testing. \textbf{\textit{tiered}Imagenet} is a larger subset of ImageNet \cite{deng2009imagenet} with $608$ classes and $779,165$ total images, which are grouped into $34$ higher-level nodes in the \textit{ImageNet} human-curated hierarchy. This set of nodes is partitioned into $20$, $6$, and $8$ disjoint sets of training, validation, and testing nodes, and the corresponding classes form the respective meta-sets. \textbf{CUB} \cite{WelinderEtal2010} dataset has a total of $200$ classes, split into training, validation and test sets following \cite{chen19closerfewshot}. We use this dataset to simulate the effect of a domain shift where the model is first trained on a ($5$-way, $1$ or $5$-shot) configuration of \textit{mini}Imagenet and then tested on the test classes of CUB, as used in \cite{chen19closerfewshot, NEURIPS2020_196f5641, DBLP:conf/icml/ZikoDGA20, NEURIPS2018_ab88b157}. 
\subsection{Implementational Details}
Let $\alpha_1$ and $\alpha_2$ respectively denote the scaling factors of the MSE and cross-entropy terms in our objective functions $\mathcal{L}_R$ and $\mathcal{L}_C$, as already defined in Subsection 4.2. The terms $\alpha$ and $\beta$ respectively denote the learning rates of the \emph{inner} and \emph{meta} updates whereas $B$ and $n$ respectively denote the number of sampled tasks and adaptation steps of the \emph{inner}-update of our end-to-end training process, as described in Algorithm 2. The hyperparameter values (\textbf{H.P.}) used for training \ourmethod{} on \textit{mini}Imagenet and \textit{tiered}Imagenet are shown in Table~\ref{tab:hyperparams}. 
\begin{table}[]
\centering
\caption{Hyperparameter values when training \ourmethod{}.}
\label{tab:hyperparams}
\resizebox{\columnwidth}{!}{%
\begin{tabular}{@{}ccccc@{}}
\toprule
                                      & \multicolumn{2}{|c|}{\cellcolor[HTML]{BEBDFF}\textbf{\textit{mini}Imagenet}}    & \multicolumn{2}{c}{\cellcolor[HTML]{FAE0C1}\textbf{\textit{tiered}Imagenet}} \\ \midrule
\multicolumn{1}{c|}{\textbf{H.P.}} &
  \multicolumn{1}{c|}{\textbf{$5$-way, $1$-shot}} &
  \multicolumn{1}{c|}{\textbf{$5$-way, $5$-shot}} &
  \multicolumn{1}{c|}{\textbf{$5$-way, $1$-shot}} &
  \textbf{$5$-way, $5$-shot} \\ \midrule
\multicolumn{1}{c|}{$\alpha_1$}       & \multicolumn{1}{c|}{1e-2} & \multicolumn{1}{c|}{1e-2} & \multicolumn{1}{c|}{1e-2}           & 1e-2           \\
\multicolumn{1}{c|}{$\alpha_2$}       & \multicolumn{1}{c|}{100}  & \multicolumn{1}{c|}{100}  & \multicolumn{1}{c|}{150}            & 150            \\
\multicolumn{1}{c|}{$\alpha$}         & \multicolumn{1}{c|}{1e-3} & \multicolumn{1}{c|}{1e-3} & \multicolumn{1}{c|}{1.5e-3}         & 1.7e-3         \\
\multicolumn{1}{c|}{\textbf{$\beta$}} & \multicolumn{1}{c|}{1e-4} & \multicolumn{1}{c|}{1e-4} & \multicolumn{1}{c|}{1.5e-4}         & 1.7e-4         \\
\multicolumn{1}{c|}{$B$}              & \multicolumn{1}{c|}{20}   & \multicolumn{1}{c|}{20}   & \multicolumn{1}{c|}{20}             & 20             \\
\multicolumn{1}{c|}{$n$}              & \multicolumn{1}{c|}{5}    & \multicolumn{1}{c|}{5}    & \multicolumn{1}{c|}{5}              & 5              \\ \bottomrule
\end{tabular}%
}
\end{table}
We apply the same hyperparameters for the cross-domain testing scenario of \textit{mini}Imagenet $\rightarrow$ CUB used for training \ourmethod{} on \textit{mini}Imagenet, for the given ($N$-way, $K$-shot) configuration. Hyperparameters are kept fixed throughout training, validation and testing for a given configuration. Adam \cite{DBLP:journals/corr/KingmaB14} optimizer is used for inner and meta-updates. Finally, the query, key and value extraction networks $f_q(,;\textbf{W}_Q)$, $f_k(.; \textbf{W}_K)$, $f_v(.; \textbf{W}_V)$ of the \ourmodule{} module only use $\texttt{Conv}_{1\times1}(.)$ and not the \texttt{LeakyReLU(0.2)} activation function for ($5$-way, $1$-shot) tasks, irrespective of the dataset. We observed that utilizing BatchNorm \cite{pmlr-v37-ioffe15} in the decoder of $z_s$ ($p_{\theta_2}$) to train \ourmethod{} on ($5$-way, $5$-shot) tasks of \textit{mini}Imagenet and on ($5$-way, $1$-shot) tasks of \textit{tiered}Imagenet leads to better scores and improved stability during training. We used the \texttt{ReLU} activation function instead of \texttt{LeakyReLU(0.2)} to carry out training on ($5$-way, $1$-shot) tasks of \textit{tiered}Imagenet. Meta-learning objectives can lead to unstable optimization processes in practice, especially when coupled with stochastic sampling in latent spaces, as also previously observed in \cite{AntoniouES19, rusu2018metalearning}. For ease of experimentation we clip the meta-gradient norm at an absolute value of 1. \ourmethod{} converges in $82,000$ and $22,500$ epochs for ($5$-way, $1$-shot) and ($5$-way, $5$-shot) tasks of \textit{mini}Imagenet, respectively and takes $67,500$ and $48,000$ epochs for convergence on ($5$-way, $1$-shot) and ($5$-way, $5$-shot) tasks of \textit{tiered}Imagenet, respectively. This translates to an average training time of $110$ hours on an $11$GB NVIDIA 1080Ti GPU. Note that we did not employ any data augmentation, feature averaging or any other data apart from the corresponding training subset $\mathcal{D}^{tr}$, during training and evaluation.
\subsection{Additional Calibration Results}
To further examine the reliability and calibration of our method, we assess the ECE, MCE \cite{pmlr-v70-guo17a} and Brier scores \cite{VERIFICATIONOFFORECASTSEXPRESSEDINTERMSOFPROBABILITY} of \ourmethod{} on the challenging \emph{cross-domain} scenario of \textit{mini}Imagenet $\rightarrow$ CUB for ($5$-way, $5$-shot) tasks. This table can be treated as an extension to Table \ref{tab:rel} since here, we compare the calibration metrics for an additional scenario. When compared against other baselines that report these metrics on the aforementioned scenario, \ourmethod{} proves to be the most calibrated with the best reliability scores. This is shown in Table~\ref{tab:rel-cross}.  
%
\begin{table}[t!]
\centering
\caption{\small Style: \textbf{best} and \underline{second best}.}
\label{tab:rel-cross}
\resizebox{0.9\columnwidth}{!}{%
\begin{tabular}{@{}lccc|@{}}
\toprule
\multicolumn{1}{l|}{\textbf{Methods}}            & \multicolumn{1}{c|}{\textbf{ECE}}   & \multicolumn{1}{c|}{\textbf{MCE}}  & \multicolumn{1}{c|}{\textbf{Brier}} \\ \midrule
\multicolumn{1}{l|}{Feature Transfer\cite{chen19closerfewshot}} & \multicolumn{1}{c|}{0.275}             & \multicolumn{1}{c|}{0.646}             & 0.772             \\
\multicolumn{1}{l|}{Baseline\cite{chen19closerfewshot}}         & \multicolumn{1}{c|}{0.315}             & \multicolumn{1}{c|}{0.537}             & 0.716             \\
\multicolumn{1}{l|}{Matching Nets\cite{NIPS2016_90e13578}}    & \multicolumn{1}{c|}{0.030}             & \multicolumn{1}{c|}{0.079}             & 0.630             \\
\multicolumn{1}{l|}{Proto Nets\cite{NIPS2017_cb8da676}}       & \multicolumn{1}{c|}{\textbf{0.009}}    & \multicolumn{1}{c|}{\underline{0.025}} & 0.604             \\
\multicolumn{1}{l|}{Relation Net\cite{Sung_2018_CVPR}}     & \multicolumn{1}{c|}{0.234}             & \multicolumn{1}{c|}{0.554}             & 0.730             \\
\multicolumn{1}{l|}{DKT+Cos\cite{DBLP:conf/nips/PatacchiolaTCOS20}}          & \multicolumn{1}{c|}{0.236}             & \multicolumn{1}{c|}{0.426}             & 0.670             \\
\multicolumn{1}{l|}{BMAML\cite{NEURIPS2018_e1021d43}}            & \multicolumn{1}{c|}{0.048}             & \multicolumn{1}{c|}{0.077}             & 0.619             \\
\multicolumn{1}{l|}{BMAML+Chaser\cite{NEURIPS2018_e1021d43}}     & \multicolumn{1}{c|}{0.066}             & \multicolumn{1}{c|}{0.260}             & 0.639             \\
\multicolumn{1}{l|}{LogSoftGP(ML)\cite{pmlr-v115-galy-fajou20a}} & \multicolumn{1}{c|}{0.220}             & \multicolumn{1}{c|}{0.513}             & 0.709             \\
\multicolumn{1}{l|}{LogSoftGP(PL)\cite{pmlr-v115-galy-fajou20a}} & \multicolumn{1}{c|}{0.022}             & \multicolumn{1}{c|}{0.042}             & 0.564             \\
\multicolumn{1}{l|}{OVE(ML)\cite{snell2021bayesian}}          & \multicolumn{1}{c|}{0.049}             & \multicolumn{1}{c|}{0.066}             & 0.576             \\
\multicolumn{1}{l|}{OVE(PL)\cite{snell2021bayesian}}          & \multicolumn{1}{c|}{\underline{0.020}} & \multicolumn{1}{c|}{0.032}             & \underline{0.556} \\ \midrule
\rowcolor{LightCyan} \multicolumn{1}{l|}{\textbf{\ourmethod{}(Ours)}} & \multicolumn{1}{c|}{\textbf{0.009}} & \multicolumn{1}{c|}{\textbf{0.02}} & \textbf{0.276} \\ \bottomrule
\end{tabular}%
}
\end{table}
\end{document}